\documentclass[sn-mathphys-num,iicol]{sn-jnl}

\usepackage{graphicx}%
\usepackage{multirow}%
\usepackage{amsmath,amssymb,amsfonts}%
\usepackage{amsthm}%
\usepackage{mathrsfs}%
\usepackage[title]{appendix}%
\usepackage{xcolor}%
\usepackage{textcomp}%
\usepackage{manyfoot}%
\usepackage{booktabs}%
\usepackage{algorithm}%
\usepackage{algorithmicx}%
\usepackage{algpseudocode}%
\usepackage{listings}%
\usepackage{bm}
\usepackage{tabularx}
\usepackage{subcaption} 
\usepackage{float}      
\usepackage{soul}
\usepackage{longtable}
\usepackage{array}
\usepackage{booktabs}

\begin{document}

\title[Article Title]{Adaptive Online Learning with LSTM Networks for Energy Price Prediction}


\author*[1]{\fnm{Salih} \sur{Salihoglu}}\email{sxs4331@miami.edu}

\author[1]{\fnm{Ibrahim} \sur{Ahmed}}\email{ibrahimahmed@miami.edu}

\author[1]{\fnm{Afshin} \sur{Asadi}}\email{axa3418@miami.edu}

\affil[1]{\orgdiv{Department of Industrial and Systems Engineering}, \orgname{University of Miami}, \orgaddress{\city{Coral Gables}, \country{FL, USA}}}


\abstract{Accurate prediction of electricity prices is crucial for stakeholders in the energy market, particularly for grid operators, energy producers, and consumers. This study focuses on developing a predictive model leveraging Long Short-Term Memory (LSTM) networks to forecast day-ahead electricity prices in the California energy market. The model incorporates a variety of features, including historical price data, weather conditions, and the energy generation mix. A novel custom loss function that integrates Mean Absolute Error (MAE), Jensen-Shannon Divergence (JSD), and a smoothness penalty is introduced to enhance the prediction accuracy and interpretability. Additionally, an adaptive online learning framework is implemented to allow the model to adapt to new data incrementally, ensuring continuous relevance and accuracy. The results demonstrate that the custom loss function can improve the model's performance, aligning predicted prices more closely with actual values, particularly during peak intervals. The adaptive online learning framework achieves the best overall performance, reducing MSE, MAE, and RMSE by approximately 23.0\%, 3.4\%, and 12.2\%, respectively, compared with the next-best model, namely the static model, which is trained once and kept fixed during testing. The inclusion of the energy generation mix further enhances the model's predictive capabilities, highlighting the importance of comprehensive feature integration. This research provides a robust framework for electricity price forecasting, which offers valuable insights and tools for better decision-making in dynamic electricity markets.}




\maketitle

\clearpage

\twocolumn[
\section*{Nomenclature}
\vspace{0.5em}
]

\begingroup
\small
\setlength{\parindent}{0pt}
\newlength{\nomlabelwidth}
\setlength{\nomlabelwidth}{1.65cm}
\newcommand{\nomentry}[2]{%
\par\noindent
\makebox[\nomlabelwidth][l]{#1}%
\parbox[t]{\dimexpr\columnwidth-\nomlabelwidth\relax}{#2}%
\par\vspace{0.25em}
}

\nomentry{\(d\)}{Index representing a day}
\nomentry{\(d'\)}{Intermediate day index used in the recurrent LSTM equations}
\nomentry{\(t\)}{Index representing an hour within a 24-hour period}
\nomentry{\(m\)}{Index representing an input feature or attribute}
\nomentry{\(i\)}{Index over entries of a probability distribution}
\nomentry{\(M\)}{Number of input features or attributes}
\nomentry{\(N\)}{Number of past days used in the sliding window}
\nomentry{\(T\)}{Number of hourly predictions in the forecast horizon; \(T=24\) in this study}
\nomentry{\(D\)}{Set of days in the training dataset}
\nomentry{\(|D|\)}{Number of days in the training dataset}
\nomentry{\(|B|\)}{Number of days or batches in the new incoming data batch set \(B\)}
\nomentry{\(|V|\)}{Number of days in the validation set \(V\)}
\nomentry{\(X_d\)}{Feature matrix for day \(d\)}
\nomentry{\(x_{d,m,t}\)}{Value of feature \(m\) at hour \(t\) of day \(d\)}
\nomentry{\(y_d\)}{Actual 24-hour electricity price vector for day \(d\)}
\nomentry{\(\hat{y}_d\)}{Predicted 24-hour electricity price vector for day \(d\)}
\nomentry{\(y_{d,t}\)}{Actual electricity price at hour \(t\) of day \(d\)}
\nomentry{\(\hat{y}_{d,t}\)}{Predicted electricity price at hour \(t\) of day \(d\)}
\nomentry{\(f_\theta(\cdot)\)}{Predictive function parameterized by \(\theta\)}
\nomentry{\(\theta\)}{Model parameter set}
\nomentry{\(\theta^*\)}{Current trained model parameter set}
\nomentry{\(\theta'\)}{Temporarily updated model parameter set}
\nomentry{\(\nabla_{\theta}\)}{Gradient with respect to the model parameter set \(\theta\)}
\nomentry{\(h^{(1)}_{d'}\)}{Hidden state of the first LSTM layer at day index \(d'\)}
\nomentry{\(h^{(2)}_{d'}\)}{Hidden state of the second LSTM layer at day index \(d'\)}
\nomentry{\(W_y\)}{Weight matrix of the dense output layer}
\nomentry{\(b_y\)}{Bias vector of the dense output layer}
\nomentry{\(\mathcal{L}(\cdot,\cdot)\)}{Pointwise prediction loss function}
\nomentry{\(L_\theta(\cdot,\cdot)\)}{Loss function evaluated under parameter set \(\theta\)}
\nomentry{\(L_V(\theta)\)}{Validation loss evaluated under parameter set \(\theta\)}
\nomentry{\(p_d\)}{Probability distribution derived from the actual price vector}
\nomentry{\(\hat{p}_d\)}{Probability distribution derived from the predicted price vector}
\nomentry{\(p_{d,t}\)}{Probability corresponding to the actual price at hour \(t\) of day \(d\)}
\nomentry{\(\hat{p}_{d,t}\)}{Probability corresponding to the predicted price at hour \(t\) of day \(d\)}
\nomentry{\(M_d\)}{Average distribution between \(p_d\) and \(\hat{p}_d\)}
\nomentry{\(KL(\cdot\|\cdot)\)}{Kullback--Leibler divergence}
\nomentry{\(JSD(\cdot\|\cdot)\)}{Jensen--Shannon divergence}
\nomentry{\(SM(\hat{y}_d)\)}{Smoothness penalty applied to the predicted price vector \(\hat{y}_d\)}
\nomentry{\(\alpha\)}{Weight assigned to the divergence component in the custom loss}
\nomentry{\(\beta\)}{Weight assigned to the smoothness penalty in the custom loss}
\nomentry{\(\eta\)}{Learning rate}
\nomentry{\(\delta\)}{Validation improvement threshold for online model updates}
\nomentry{\(S\)}{Original training dataset collected so far}
\nomentry{\(V\)}{Validation set used for update decisions}
\nomentry{\(B\)}{New incoming data batch set}
\nomentry{\(b\)}{Index representing a new incoming batch or day, where \(b \in B\)}
\nomentry{\(X_B, y_B\)}{Feature and target values for the new incoming batch set \(B\)}
\nomentry{\(X_V, y_V\)}{Feature and target values for the validation set \(V\)}

\endgroup

\clearpage

\section{Introduction}
\label{intro}

In recent years, the accurate prediction of electricity prices for deregulated markets has become increasingly crucial for various stakeholders in the energy market, including grid operators, energy producers, and consumers. This need is particularly pronounced in regions like California, where dynamic pricing is employed. Dynamic pricing adjusts electricity prices in real-time based on supply and demand fluctuations, encouraging consumers to modify their usage patterns and thus ensuring a balanced and efficient power grid. For grid operators, it aids in demand response management, balancing supply and demand, and reducing operational costs. For energy producers, it facilitates optimal bidding strategies in the electricity market. Consumers, on the other hand, benefit from better-informed decisions about their energy usage, leading to cost savings and enhanced energy efficiency. However, it is known that creating effective price prediction models is challenging due to high-frequency fluctuations and price volatility in the electricity market \citep{Ghimire2024}.

The inherent volatility and non-linearity of electricity prices, driven by dynamic factors such as weather conditions, demand patterns, and generation mix, pose significant challenges to traditional forecasting methods. The traditional statistical models often fall short in capturing the complex and dynamic nature of electricity price movements, leading to inaccurate predictions and sub-optimal decision-making. To address these challenges, advanced machine learning techniques have emerged as powerful tools for time series forecasting. Among these advanced techniques, Long Short-Term Memory (LSTM) networks, a special kind of recurrent neural network (RNN), have shown great promise (see for instance \citealp{sagheer2019time, salihoglu2024enhancing, dubey2021study}). LSTMs are designed to capture long-term dependencies and temporal patterns in sequential data, making them well-suited for applications like electricity price prediction. The ability of LSTMs to remember information for long periods helps in modeling the price dynamics more accurately than traditional methods.

This study focuses on developing an LSTM-based model to predict electricity prices in California, leveraging historical price data and a variety of related features. The primary objective is to design a robust prediction system to forecast the prices for the next 24 hours. This approach ensures that the model can adapt to the dynamic nature of electricity markets and provide accurate and timely predictions, which are essential for both operational planning and strategic decision-making. The methodology adopted in this article involves a three-step mathematical process. First, we formulate the problem as a 24-vector prediction task through a recurrent neural network-based prediction model. Second, we introduce a custom loss function tailored to the specific requirements of electricity price forecasting. This custom loss function combines the multi-dimensional regression loss with additional components to account for prediction smoothness and distributional divergence, enhancing the robustness of the predictions. Finally, we implement an adaptive online learning approach, allowing the model to update its parameters incrementally as new data becomes available. This incremental learning capability ensures that the model remains relevant and accurate over time, adapting to changes in market conditions. In summary, this study aims to bridge the gap between advanced machine learning techniques and practical electricity price forecasting. By leveraging LSTM networks and incorporating custom loss functions and online learning, the proposed model aspires to deliver high accuracy and robustness, providing stakeholders with a reliable tool for navigating the complexities of dynamic electricity markets. We will also discuss how an additional set of features can improve the prediction accuracy of the model and their effectiveness in doing so. Although online learning has been explored for energy forecasting, many existing approaches rely on continuous or periodic model updates without explicitly evaluating whether newly arriving data improves generalization performance. As a result, such methods may become sensitive to noisy, non-representative, or short-term anomalous observations in volatile electricity markets. To address this limitation, this study proposes a unified framework that combines a custom loss function for capturing pointwise accuracy, distributional similarity, and temporal smoothness with a validation-based selective update mechanism, which allows the model to adapt only when new data provides meaningful predictive improvement. The primary contributions of this work are summarized as follows:
    
\begin{itemize}
\setlength{\leftskip}{1.5em}
    \item[(i)] Day-ahead electricity price forecasting is formulated as a 24-dimensional vector prediction problem, which allows intra-day temporal dependencies to be explicitly learned rather than relying on recursive single-step forecasts.
    \item[(ii)] A novel custom loss function is proposed that combines pointwise error minimization with distributional similarity (via Jensen–Shannon Divergence) and temporal smoothness constraints, which leads to more realistic and stable price trajectories.
    \item[(iii)] An adaptive online learning framework with validation-based selective updates is introduced, which improves robustness to noisy or non-representative data compared to conventional rolling or periodic retraining strategies.
    \item[(iv)] A systematic evaluation of the impact of exogenous variables, including weather conditions and generation mix data, is conducted under both static and adaptive learning settings.
\end{itemize}

The rest of this article is organized as follows: Section 2 provides a comprehensive background and literature review, detailing the importance of energy pricing and discussing existing methodologies and their shortcomings. Section 3 introduces the main prediction framework, defines the day-ahead energy price prediction problem, describes the model architecture, and details the development of a custom loss function designed to enhance interpretability and robustness. Section 4 explains the online learning approach, detailing the process of incremental model updates and decision-making for incorporating new data. Section 5 discusses the experimental results based on a real dataset, including data description, preprocessing steps, and a detailed analysis of the model's performance under different configurations. Finally, Section 6 concludes the article with a summary of findings, limitations, and suggestions for future work.

\section{Literature Review and Background}
\label{chp:b2}

Dynamic energy pricing is crucial for balancing supply and demand in electricity markets, ensuring grid stability, and promoting efficient energy use. It involves setting electricity costs that fluctuate based on real-time factors like production expenses, market demand, and regulatory policies. Common models include Time-of-Use (TOU) and Real-Time Pricing (RTP), which adjust prices according to the time of day and current market conditions. Accurate dynamic pricing signals consumers to modify their usage patterns, supports investment in renewable energy, and aids in optimal grid operation, thereby preventing overloads and blackouts. It also helps producers make informed decisions about generation and market bidding. Effective dynamic energy pricing mechanisms are essential for achieving economic efficiency, energy security, and environmental sustainability in the power sector. However, accurate electricity price forecasting (EPF), particularly for day-ahead markets, is highly challenging due to factors such as the high penetration of renewable energy, variations in load, geographic differences, and interactions between price zones \citep{Meng2024}. 

\subsection{Statistical and Time-Series Forecasting Models}
Various statistical models have been used for electricity price forecasting. \citet{garcia2005garch} applied a GARCH model to hourly electricity prices from Spain and California, showing improved performance over ARIMA models in capturing volatility and price spikes, although with limited ability to model nonlinear relationships and long-term dependencies. \citet{hickey2012forecasting} compared ARMAX and several GARCH-type models across five MISO hubs and found that APARCH models performed well in deregulated markets, while simpler GARCH models were more suitable for regulated settings. \citet{kosater2006can} used Markov regime-switching models for hourly spot price forecasting and showed that nonlinear regime-based models can outperform linear autoregressive specifications, particularly for long-run forecasts. Overall, these studies demonstrate the usefulness of statistical models for capturing volatility and market regimes, but also highlight their limitations in modeling complex nonlinear and temporal patterns in electricity prices. 

Time-series models, particularly ARIMA-based approaches, have been widely used in Locational Marginal Price (LMP) forecasting due to their interpretability. For example, \cite{contreras2004arima} and \cite{zhou2006electricity} applied ARIMA models to generate LMP scenarios and estimate confidence intervals. However, conventional scalar time-series models often struggle to capture intra-day structure, nonstationarity, and multiple seasonal patterns in electricity prices. To address these issues, functional time-series approaches have modeled daily 24-hour profiles as continuous curves. \citet{jan2022short} showed that functional autoregressive models outperform traditional autoregressive and naïve benchmarks in short-term electricity price forecasting, while \citet{jan2025forecasting} demonstrated that incorporating exogenous variables within a functional data framework improves day-ahead price forecasts over classical AR, ARX, ARIMA, and ARIMAX models. Similar functional approaches have also improved electricity demand forecasting, as shown by \citet{jan2025advancing} and \citet{shah2022functional}.

\subsection{Deep Learning and Hybrid Forecasting Models}

Recently, with the emergence of various machine learning techniques, researchers have increasingly shifted their focus toward leveraging these methods for forecasting tasks due to their ability to handle complex correlations and nonlinear dependencies. In the field of electricity price forecasting, deep learning methods have garnered particular attention for modeling highly nonlinear temporal dynamics. Among these methods, LSTM networks have shown considerable promise and have been successfully employed in several studies to forecast day-ahead electricity prices. For instance, \citet{jiang2018day} employed an LSTM-based model to predict day-ahead electricity prices for Australia and Singapore using historical prices, weather conditions, system demand, oil prices, and temporal features, achieving improved performance through stacked LSTM architectures and recursive multi-hour forecasting. Similarly, \citet{zhou2019optimized} optimized LSTM networks for the Pennsylvania–New Jersey–Maryland (PJM) market, demonstrating enhanced predictive accuracy through heterogeneous architectures and hyperparameter optimization.

Other studies have explored hybrid models combining LSTM with other techniques to further enhance predictive performance. \citet{kuo2018electricity} proposes a model for electricity price forecasting, which uses historical prices, loads, and factors like climate and market demands, leveraging data from the PJM Regulation Zone Preliminary Billing Data. The EPNet, a hybrid model combining Convolutional Neural Networks (CNN) and LSTM networks, outperforms traditional models such as Support Vector Machine (SVM), Random Forest (RF), Decision Tree (DT), Multilayer Perceptron (MLP), standalone CNN, and LSTM. Traditional models struggle with high errors and poor trend prediction, often needing extra feature selection and calculations. EPNet's hybrid approach enhances accuracy and practical applicability, aiding power generators and consumers in decision-making. \citet{zhang2020adaptive} presents a novel approach combining variational mode decomposition (VMD), self-adaptive particle swarm optimization (SAPSO), seasonal autoregressive integrated moving average (SARIMA), and deep belief network (DBN) for accurate short-term electricity price forecasting. The model is validated using data from the Australian, PJM, and Spanish electricity markets. The proposed hybrid model demonstrates superior performance in both normal price and price spike forecasting compared to traditional models.

\subsection{Online and Adaptive Learning Approaches}

Online learning is crucial in dynamic environments such as energy markets, where prediction models must adapt to changing market conditions and evolving data distributions. Several studies have explored online and adaptive learning techniques for energy-related forecasting applications. \citet{kim2023online} developed an online machine learning approach for system marginal price forecasting using multiple economic indicators and combined batch learning with online updates. \citet{wang2024reinforcement} proposed a reinforcement-learning-enhanced online learning strategy for short-term load forecasting, where an agent dynamically selects training data based on recent prediction performance and demand variability. \citet{nayak2025probabilistic} introduced a probabilistic online learning framework for short-term wind power forecasting using online ensemble bagging regression and sequential gradient-based updates. \citet{melgar2025online} proposed StreamWNN, an incremental nearest-neighbor-based online learning algorithm for real-time forecasting of electricity demand and electricity prices. \citet{melgar2023novel} developed a distributed forecasting method based on information fusion and incremental learning for streaming time series. \citet{carnero2024online} proposed a Kafka-based machine learning framework to support continuous online training and automatic model upgrading in streaming environments. \citet{arshad2025performance} further showed that data imbalance can sharply degrade the performance of learning models, emphasizing the need to account for distributional shifts and evaluation sensitivity in adaptive forecasting.

Recent studies on electrical load forecasting have further emphasized that forecasting models deployed in smart-grid environments may suffer from concept drift due to changes in data distributions, input features, operating conditions, and generation modalities. Azeem et al. discussed concept-drift scenarios in online electrical load forecasting and highlighted that static models may deteriorate when electrical data streams evolve over time \citep{azeem2022concept}. Similarly, Azeem et al. showed that conventional ARIMA, ANN, and LSTM models can experience performance degradation under parameter and modality changes in heterogeneous smart-grid environments \citep{azeem2022deterioration}. More recent work by Azeem et al. proposed an adaptive ensemble LSTM framework with dynamic feature engineering and attention mechanisms to improve load forecasting under concept drift \citep{azeem2025mitigating}. In a related study, Azeem et al. developed a transfer-learning-enabled adaptive LSTM framework for load forecasting across different generation modalities \citep{azeem2024transfer}.

Recent renewable-energy forecasting studies have also addressed distribution shifts across heterogeneous data sources. Passalis et al. proposed a residual adaptive input normalization approach for renewable energy generation forecasting across multiple countries \citep{passalis2025residual}. Dimitriadis et al. developed a deep learning framework for photovoltaic power forecasting across interconnected countries using foundation training, country-specific fine-tuning, and recalibration \citep{dimitriadis2025deep}. These findings further motivate the need for adaptive forecasting frameworks that can update selectively as new data becomes available, rather than relying solely on fixed offline models or unconditional periodic retraining.

\subsection{Exogenous Variables and Generation Mix Data}

While existing literature has explored various models for LMP forecasting, the incorporation of additional parameters such as daily weather information and fuel mix data presents an opportunity to enhance predictive accuracy and robustness.
Weather conditions play a significant role in electricity demand and generation patterns, thus exerting influence on LMPs. Incorporating daily weather information into forecasting models enables capturing the impact of temperature, humidity, wind speed, and solar radiation on electricity demand and generation patterns \citep{neumann2023using}. \citet{son2017short} and \citet{wang2019daily} demonstrated improved forecasting accuracy with weather data integration. The composition of the energy generation mix, encompassing sources such as coal, natural gas, nuclear, renewables, and battery storage, significantly influences LMPs due to variations in production costs, availability, intermittency and environmental regulations. Integrating fuel mix information into forecasting models enables capturing the interplay between different energy sources and their respective contributions to the overall market dynamics. \citet{tschora2022electricity} and \citet{Thesis} emphasized the importance of renewable generation information for comprehensive forecasting models. This is also consistent with prior work showing that load forecasting performance depends on the availability and selection of relevant meteorological, operational, and generation-related variables, particularly in industrial and smart-grid environments with diverse energy sources \citep{azeem2021implication}.

\citet{8894352} exploited the energy mix data provided by the New York Independent System Operator (NYISO) to forecast day-ahead load and LMP and optimize energy storage scheduling while considering the volatility of LMP due to charging and discharging of the energy storage. However, this model did not expand to include the impact of other energy generation components. Wind energy has been particularly studied as one important component of the generation mix. \citet{6575188} explored the impact of energy component dynamics on LMP forecasting, highlighting the importance of incorporating wind generation information for comprehensive price forecasting models. Furthermore, \citet{5512565} performed a simulation to analyze the impact of increasing wind power integration on LMP, where they found a correlation between increased penetration of wind energy and average LMP. Still, research focusing on fuel mix data integration in LMP forecasting is relatively scarce but holds promise for enhancing predictive accuracy.

To clarify the methodological distinction between the proposed framework and the reviewed literature, Table~\ref{tab:summary_positioning} summarizes the main research streams and highlights the gap addressed by this study.

\begin{table*}[!t]
\centering
\small
\caption{Summary comparison of related research streams and positioning of the proposed framework}
\label{tab:summary_positioning}
\begin{tabularx}{\textwidth}{@{}p{3.0cm}p{4.2cm}X@{}}
\toprule
Research stream & Representative studies & Main gap addressed by this study \\
\midrule

Statistical and time-series models 
& \citet{garcia2005garch,hickey2012forecasting,kosater2006can,contreras2004arima,zhou2006electricity} 
& These methods provide interpretable baselines for volatility and LMP forecasting, but are limited in capturing nonlinear 24-hour price trajectories and adapting to evolving market conditions. \\
\addlinespace[0.6em]

Functional time-series models 
& \citet{jan2022short,jan2025forecasting,jan2025advancing,shah2022functional} 
& These approaches capture daily profile structures, but generally do not provide validation-based selective online updates for changing market conditions. \\
\addlinespace[0.6em]

Deep learning and hybrid models 
& \citet{jiang2018day,zhou2019optimized,kuo2018electricity,zhang2020adaptive} 
& These models improve nonlinear forecasting performance, but often rely on static training and do not explicitly combine 24-hour trajectory regularization with selective online adaptation. \\
\addlinespace[0.6em]

Online/adaptive and concept-drift methods 
& \citet{kim2023online,wang2024reinforcement,nayak2025probabilistic,melgar2025online,melgar2023novel,carnero2024online,arshad2025performance,azeem2022concept,azeem2022deterioration,azeem2025mitigating,azeem2024transfer,passalis2025residual,dimitriadis2025deep} 
& These studies address online learning, concept drift, and distribution shift, but many focus on load or renewable-generation forecasting, or apply continuous updates without validation-based acceptance of new data. \\
\addlinespace[0.6em]

Exogenous and generation-mix studies 
& \citet{neumann2023using,son2017short,wang2019daily,tschora2022electricity,Thesis,azeem2021implication,8894352,6575188,5512565} 
& These studies support the value of weather and generation-related inputs, but do not combine such features with custom trajectory-level loss regularization and selective online learning for EPF. \\
\addlinespace[0.6em]

Proposed framework 
& This study 
& Combines 24-hour LSTM price-vector forecasting, MAE--JSD--smoothness loss, weather and generation-mix features, and validation-based selective online model updates. \\

\bottomrule
\end{tabularx}
\end{table*}

In summary, the reviewed studies show substantial progress in electricity price forecasting through statistical models, functional time-series approaches, deep learning, hybrid models, online learning, and exogenous feature integration. However, as summarized in Table~\ref{tab:summary_positioning}, most existing approaches either focus on static offline forecasting, apply continuous updates without explicitly validating whether new data improve generalization, or address load and renewable-generation forecasting rather than day-ahead electricity price forecasting. In contrast, the proposed framework combines 24-hour LSTM price-vector forecasting, a MAE--JSD--smoothness custom loss, weather and generation-mix inputs, and validation-based selective online model updating. These observations motivate the methodological framework presented in the next section.

\section{The Main Prediction Framework}
\label{proposedmethod}

In this section, the structure of the prediction framework is discussed. First, we provide a formal definition of a day-ahead energy price prediction problem. In the context of electricity markets, the day-ahead prediction problem involves forecasting the electricity prices for the next 24 hours based on historical data and other relevant features as discussed below. 

\noindent
- \textbf{A Day-Ahead Energy Price Prediction}: 
Let us define a time series of $M$ features or attributes collected over the 24 hours of day $d$ as $\mathbf{X}_{d}$, that is,
\begin{equation}
\mathbf{X}_{d} = \begin{bmatrix}
x_{d,1,1} & x_{d,1,2} & \cdots & x_{d,1,24} \\
x_{d,2,1} & x_{d,2,2} & \cdots & x_{d,2,24} \\
\vdots & \vdots & \ddots & \vdots \\
x_{d,M,1} & x_{d,M,2} & \cdots & x_{d,M,24}
\end{bmatrix}_{M\times24},    
\end{equation}
where $x_{d,m,t}$ represents the $m$-th feature at time $t$ of day $d$. Now, the goal is to predict the vector of electricity prices $\hat{\mathbf{y}}_{d+1}= [\hat{y}_{d+1,1},\hat{y}_{d+1,2},...,\hat{y}_{d+1,24}]$ for the next 24 hours. The objective is to learn a predictive function $f$ that maps a sequence of the last $N$ days of features to future prices using a sliding window approach. That is, 

\begin{equation}
\begin{aligned}
\hat{\mathbf{y}}_{d+1}
&= f_{\bm{\theta}}\big(
\mathbf{X}_{d-N+1}, \ldots,
\mathbf{X}_{d}, \mathbf{X}_{d+1}
\big) \\
&= f_{\bm{\theta}}\big(\mathbf{X}_{d-N+1:d+1}\big),
\end{aligned}
\end{equation}
where $\hat{\mathbf{y}}_{d+1}$ is the predicted price vector for the next 24 hours (day $d+1$) and $N$ is the number of past days used for making the prediction. In the context of energy price prediction, since it is often possible to forecast the features for the next 24 hours, we can incorporate these predicted features into the model to enhance the accuracy of future price predictions. The window size $N$ is also a critical parameter in this model. A larger window size $N$ allows the model to capture longer temporal dependencies but may increase the complexity and computational load. Conversely, a smaller $N$ may miss important patterns regarding the dynamic nature of the market. 

The non-stationary predictive function $f_{\bm{\theta}}$ (with characteristic parameters defined by set $\bm{\theta}$) for day-ahead energy price prediction can be chosen depending on its ability to capture complex temporal dependencies and nonlinear relationships in time series data. Common models include LSTM networks, RNNs, Convolutional Neural Networks (CNNs), transformer models, and hybrid models. The predictive function $f_{\bm{\theta}}$ should have essential properties to ensure accurate predictions. For instance, it needs to capture temporal dependencies, model nonlinear relationships, and be computationally efficient and scalable to manage large datasets. Additionally, $f_{\bm{\theta}}$ should be robust against noise and outliers in the data, providing reliable predictions even when the input data is imperfect. Finally, the model must generalize well to new, unseen data, avoiding overfitting while maintaining high predictive accuracy. The function $f_{\bm{\theta}}$ in this article is modeled by an LSTM network. The proposed methodology integrates a holistic approach to forecasting electricity prices, leveraging a rich dataset that encompasses a wide range of features. 
\subsection{Model Architecture}
\begin{figure}[!t]
 \centering
 \includegraphics[width=\columnwidth]{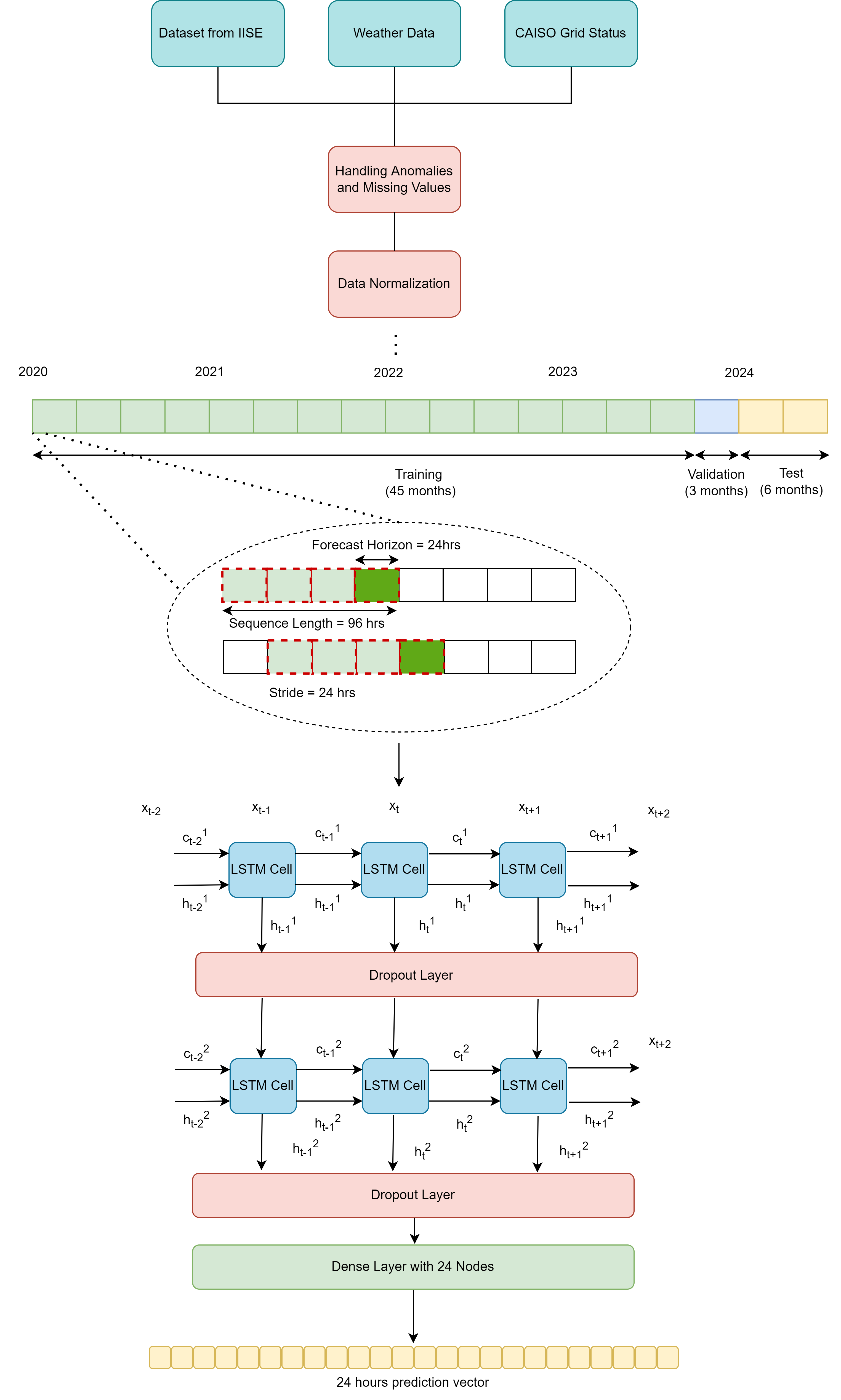}
 \caption{An overview of the model architecture}
 \label{fig:architecture}
\end{figure}

For energy price prediction, the model takes a sequence of the last $N$ days of features. This input sequence is fed into the LSTM layers, which process the data to capture temporal relationships. The output of the LSTM layers is then passed to a dense layer, which produces the 24-hour ahead electricity price predictions. After evaluating various model architectures, the following LSTM configuration, illustrated in Figure 1, emerged as the most effective one for forecasting electricity prices.
Given an input sequence $\mathbf{X}_{d-N+1}, \mathbf{X}_{d-N+2}, \ldots, \mathbf{X}_{d+1}$, the LSTM layers and dense layer in the neural network transform this input into the predicted output sequence
$\hat{\mathbf{y}}_{d+1,1:24}$, where
$\hat{\mathbf{y}}_{d+1,1:24} \equiv \hat{\mathbf{y}}_{d+1}$ denotes the 24-hour predicted electricity price vector for day $d+1$. The transformation can be represented as follows:
\begin{equation}
\begin{aligned}
\begin{array}{lll}
\mathbf{h}_{d'}^{(1)}
&= \text{LSTM}_1\big(\mathbf{X}_{d'}, \mathbf{h}_{d'-1}^{(1)}\big), \\
\mathbf{h}_{d'}^{(2)}
&= \text{LSTM}_2\big(\mathbf{h}_{d'}^{(1)}, \mathbf{h}_{d'-1}^{(2)}\big), \\
\hat{\mathbf{y}}_{d+1,1:24}
&= \text{Dense}\big(\mathbf{h}_{d'}^{(2)}\big),
\end{array}
\end{aligned}
\label{eq:lstm_transformation}
\end{equation}
where \(d'=d-N+1,\ldots,d+1\).
Here, $\text{LSTM}_1$ and $\text{LSTM}_2$ represent the first and second LSTM layers, respectively. The recurrent layers $\mathbf{h}_{d'}^{(1)}$ and $\mathbf{h}_{d'}^{(2)}$ are the hidden states of the first and second LSTM layers. Finally, the $\text{Dense}$ function represents the dense layer that outputs the predicted 24-hour ahead electricity prices as follows: 
\begin{equation}
\hat{\mathbf{y}}_{d+1,1:24} = \mathbf{W}_y \mathbf{h}_{d'}^{(2)} + \mathbf{b}_y,
\end{equation}
where $\mathbf{W}_y$ is the weight matrix of the dense layer with shape $(24, \text{hidden\_size})$, 
$\mathbf{h}_{d'}^{(2)}$ is the final hidden state from the second LSTM layer with shape $(\text{hidden\_size}, 1)$, and $\mathbf{b}_y$ is the bias vector of the dense layer with shape $(24, 1)$. Thus, the predicted prices $\hat{\mathbf{y}}_{d+1}$ are derived from the input sequence $\mathbf{X}_{d-N+1}, \mathbf{X}_{d-N+2}$, \ldots, $\mathbf{X}_{d+1}$ by processing through the LSTM layers and the dense layer. The model architecture includes two LSTM layers, each with 256 units. The first LSTM layer has a dropout rate of 0.3 to capture temporal dependencies within the input sequence, while the second LSTM layer, also with 256 units, is followed by another dropout layer with a rate of 0.3 to prevent overfitting. The output layer is a dense layer with linear activation, designed to predict the 24-hour ahead electricity prices by leveraging the sequential input to produce a vector of hourly prices. The model's forecasting capability was rigorously assessed using Root Mean Squared Error (RMSE), Mean Squared Error (MSE), and Mean Absolute Error (MAE) metrics, prioritizing accuracy and reliability in our predictions. Figure \ref{fig:architecture} shows the structure of the model and all the key steps.
To ensure the robustness of our LSTM model and optimize its performance, we employed hyperparameter tuning as a critical step in our framework. We explored a range of configurations for the model's architecture and training process, including variations in the number of LSTM units, dropout rates, and learning rates. The fine-tuned model, with specific settings for these hyperparameters, yielded better results.

For notational simplicity in the training objective, we re-index the target day as \(d\), so that \(\hat{\mathbf{y}}_d\) denotes the prediction for the day being forecast.
To train the model, we use a supervised learning approach. The training data consists of \(|D|\) days of historical data. Given the training data 
\(\{(\mathbf{X}_{d}, \mathbf{y}_{d})\}_{d\in D}\), 
the LSTM model parameters are optimized by minimizing a loss function that measures the difference between the predicted 24-hour price vector \(\hat{\mathbf{y}}_{d,1:24} \equiv \hat{\mathbf{y}}_{d}\) and the actual 24-hour price vector \(\mathbf{y}_{d,1:24} \equiv \mathbf{y}_{d}\) over all days in the training period. The loss function can be formulated as follows:
\begin{equation}
\text{Loss}
= \frac{1}{24|D|}
\sum_{d \in D} \sum_{t=1}^{24}
\mathcal{L}(\hat{y}_{d,t}, y_{d,t}),
\end{equation}
where
\(\hat{\mathbf{y}}_{d,1:24}
= f_{\bm{\theta}}\big(\mathbf{X}_{d-N+1:d}\big)\),
\(D\) is the set of days in the training data and \(|D|\) is the total number of days,
\(\hat{y}_{d,t}\) is the predicted price for the \(t\)-th hour of the \(d\)-th day,
\(y_{d,t}\) is the actual price for the \(t\)-th hour of the \(d\)-th day, and
\(\mathcal{L}\) is the loss function, such as MSE.
\subsection{Custom Loss Function to Improve the Prediction Interpretability and Robustness}
In the context of predicting electricity prices, it is crucial to have a loss function that not only minimizes the prediction error but also ensures the stability and reliability of the predictions. The custom loss function used in this study is designed to achieve these goals by incorporating regularizers in the form of custom loss functions as discussed below. 
\noindent
\textbf{- Distribution Similarity}: When using an LSTM network to predict 24-hour intervals of energy prices, it is crucial that the model captures specific trends observed in the original data, such as peaks, lows, and high-price intervals. To address this, the model's training process must be influenced to accurately reflect these trends in its predictions. To ensure that the predicted price distribution closely matches the actual distribution in each 24-hour interval, we incorporate the Jensen-Shannon Divergence (JSD) component into the loss function. JSD is a symmetric and smoothed version of the Kullback-Leibler (KL) divergence, which measures the similarity between two probability distributions. By including JSD in the loss function, the model is penalized when the predicted distribution deviates from the actual distribution, thereby encouraging the model to produce predictions that more faithfully replicate the observed trends in the data. The Kullback-Leibler divergence $ \text{KL}(\mathbf{p}_d \parallel \mathbf{M}_d) $ is given by:
\begin{equation}
\operatorname{KL}(\mathbf{p}_d \parallel \mathbf{M}_d)
= \sum_i \mathbf{p}_d(i)
\log\left(
\frac{\mathbf{p}_d(i)}{\mathbf{M}_d(i)}
\right).
\end{equation}
where $ \mathbf{p}_d $ and $ \hat{\mathbf{p}}_d $ are the true and predicted probability distributions, respectively, and \(\mathbf{M}_d = \frac{1}{2} (\hat{\mathbf{p}}_d + \mathbf{p}_d)\) is the average distribution. The index \( i \) iterates over each hour in the 24-hour prediction interval. The JSD term ensures that the predicted price distribution remains close to the actual distribution, reducing the likelihood of outlier predictions. Given the predicted values \(\hat{\mathbf{y}}_d = [\hat{y}_{d,1}, \ldots, \hat{y}_{d,24}]\) and actual values \(\mathbf{y}_d = [y_{d,1}, \ldots, y_{d,24}]\) for day $d$, we first need to convert these into probability distributions using the softmax function:
\begin{equation}
\begin{aligned}
\hat{p}_{d,t}
&= \frac{\exp(\hat{y}_{d,t})}
{\sum_{\tau=1}^{24} \exp(\hat{y}_{d,\tau})}, \\
p_{d,t}
&= \frac{\exp(y_{d,t})}
{\sum_{\tau=1}^{24} \exp(y_{d,\tau})},
\end{aligned}
\end{equation}
for all \(d \in D\) and \(t \in \{1,\ldots,24\}\),
where \(\hat{p}_{d,t}\) and \(p_{d,t}\) are the probabilities corresponding to the predicted and actual values for time step \(t\) within day \(d\). The softmax transformation is used here to convert each 24-hour price vector into a normalized intra-day profile for measuring distributional similarity. Since electricity prices may contain negative values and large spikes, the JSD component is not intended to capture absolute price levels. Rather, it acts as a shape-based regularizer that encourages the predicted profile to preserve the relative intra-day structure of the actual price trajectory, including peak and low-price intervals. Absolute magnitude errors are still directly penalized by the MAE term. Thus, the JSD component is used only as a complementary regularizer rather than as a substitute for price-level accuracy. To add JSD to the loss function, we compute the JSD for each 24-hour interval between the predicted (\(\hat{\mathbf{y}}_d\)) and actual (\(\mathbf{y}_d\)) distributions, and add it to the original loss function. The JSD loss function becomes:
\begin{equation}
\begin{aligned}
\text{JSD}(\hat{\mathbf{p}}_d \parallel \mathbf{p}_d)
= \frac{1}{2} \Big[
&\text{KL}(\hat{\mathbf{p}}_d \parallel \mathbf{M}_d) \\
&+ \text{KL}(\mathbf{p}_d \parallel \mathbf{M}_d)
\Big]
\end{aligned}
\end{equation}
The above JSD loss term when added to the main loss function ensures that the LSTM model not only minimizes the prediction error but also maintains the distributional characteristics of the actual energy prices.

\noindent
\textbf{- Smoothness}: To promote smooth transitions between consecutive predictions, a smoothness penalty is included in the loss function. This penalty helps to avoid abrupt changes in predictions, which are unrealistic in the context of electricity prices. The smoothness penalty is defined as:
\begin{equation}
\text{SM}(\hat{\mathbf{y}}_{d}) = \sum_{t=1}^{T-1} \left( \hat{y}_{d,t+1} - \hat{y}_{d,t} \right)^2,
\end{equation}
where $ \hat{y}_{d,t} $ and $ \hat{y}_{d,t+1} $ are consecutive predicted values, and $T$ is the total number of predictions (i.e., $T=24$ in this article).

\noindent
\textbf{- Modified Custom Loss Function}: The custom loss function after taking into account the distribution similarity and smoothness terms is now a weighted combination of the MAE, JSD, and smoothness penalty, that is
\begin{equation}
\operatorname{Loss}
= \operatorname{MAE}
+ \alpha \operatorname{JSD}
+ \beta \operatorname{SM},
\end{equation}
where $ \alpha $ and $ \beta $ are hyperparameters (that need to be fine-tuned) that control the contribution of the JSD and smoothness penalty, respectively. The final custom loss function used in this study is defined as:
\begin{equation}
\begin{aligned}
\text{Loss}
&= \frac{1}{24|D|}
\sum_{d \in D} \sum_{t=1}^{24}
\mathcal{L}(\hat{y}_{d,t}, y_{d,t}) \\
&\quad + \alpha \sum_{d \in D}
\operatorname{JSD}(\hat{\mathbf{p}}_d \parallel \mathbf{p}_d) \\
&\quad + \beta \sum_{d \in D}
\operatorname{SM}(\hat{\mathbf{y}}_{d}).
\end{aligned}
\end{equation}
In summary, this custom loss function enhances the interpretability and robustness of electricity price predictions by combining three complementary components: the MAE term minimizes pointwise prediction errors, the JSD term aligns the predicted intra-day profile with the actual profile, and the smoothness penalty encourages stable transitions between consecutive hourly predictions. This comprehensive approach leads to more reliable and stable energy price prediction.

\section{Adaptive Online Learning}

The incorporation of real-time data is imperative due to the market's highly dynamic and volatile nature for energy price prediction. Factors such as supply-demand imbalances, weather conditions, regulatory changes, and geopolitical events can cause rapid and unpredictable fluctuations in energy prices. The high frequency and variability of this data render traditional static models, which update periodically, insufficient and prone to sub-optimal performance. To mitigate these issues, online learning, or incremental learning, is a highly effective approach. Online learning algorithms are designed to update model parameters incrementally with each new data point, rather than relying on bulk data processing. This allows the model to rapidly adapt to the latest trends and changes, ensuring it remains accurate and relevant in a continuously evolving environment. By adopting online learning, we can develop a robust, adaptive, and responsive energy price prediction system capable of keeping pace with the rapid changes in the energy market, thereby delivering more accurate and timely predictions. This section describes the online learning approach used in this study, detailing the forward propagation, loss computation, backpropagation, parameter update processes, and the decision-making process for including new data in the training set. 

In this article, we propose an adaptive online learning framework, which evaluates the relevance and distribution of new data to decide whether it should be used for updating the model. Based on this approach, the model dynamically adjusts its parameters based on the feedback from incoming data and a validation set. The core idea is to ensure that updates to the model improve its performance on a validation set, thereby maintaining or enhancing the model's generalization capabilities. Also, the model is updated incrementally as new data arrives, rather than retraining from scratch. The validation-based approach ensures that only beneficial updates are applied, making the model more robust to noisy or non-representative data. The model can adapt to changes in the data distribution over time, making it suitable for the dynamic environment in the energy market where data characteristics evolve. Let us denote $S$ as the set of days in the original training dataset collected so far, $V$ as the set of days in the validation set, and $B$ as the set of days in the new dataset. In a most simple case, $V$ and $B$ can contain only the data for one day. Now, given a new batch of data ($b \in B$) consisting of features $ \bm{X}_{b} $ and corresponding target values $ \bm{y}_{b} $, the online learning model operates as follows:

\noindent \textbf{I. Predictions Step}: estimate $ \hat{\bm{y}}_{b} $ through forward propagation given the current model with parameter value $\bm{\theta}^*$ for all new datasets as 
\begin{equation}
\hat{y}_{b} = f_{\bm{\theta}^*}(\bm{X}_{b-N+1:b}), \forall b \in B,
\end{equation}
where $ f_{\bm{\theta}^*} $ represents the neural network model with parameters $ \bm{\theta^*} $ obtained so far.

\noindent
\textbf{II. Loss Evaluation}: calculate the loss value for the new data as follows:

\begin{equation}
\begin{aligned}
L_{\bm{\theta}^*}(\bm{X}_{B}, \bm{y}_{B})
&= \frac{1}{24|B|}
\sum_{b \in B} \sum_{t=1}^{24}
\mathcal{L}(\hat{y}_{b,t}, y_{b,t}) \\
&\quad + \alpha \sum_{b \in B}
\operatorname{JSD}(\hat{\mathbf{p}}_b \parallel \mathbf{p}_b) \\
&\quad + \beta \sum_{b \in B}
\operatorname{SM}(\hat{\mathbf{y}}_{b}).
\end{aligned}
\end{equation}
\noindent
\textbf{III. Temporary Parameter Update}:
estimate update parameters based on gradient calculation
\begin{equation}\label{eq:gra}
 \bm{\theta}' = \bm{\theta}^* - \eta 
 \nabla_{\bm{\theta}^*} L_{\bm{\theta}^*}(\bm{X}_{B}, \bm{y}_{B}).
\end{equation}
where $\eta$ is the learning rate, which determines the step size in the direction of the gradient. A smaller $\eta$
means smaller updates, leading to slower but more stable learning, while a larger $\eta$ means larger updates, which can speed up learning but might cause instability.

\noindent
\textbf{IV. Validation Set Forward Pass}: Evaluate the loss on the validation set using both the current and proposed parameters, that is to calculate $L_{\bm{\theta}^*}(\bm{X}_{V}, \bm{y}_{V})$ and $L_{\bm{\theta}'}(\bm{X}_{V}, \bm{y}_{V})$ from Eq. (7).

\begin{figure*}[!t]
 \centering
 \includegraphics[width=0.95\textwidth]{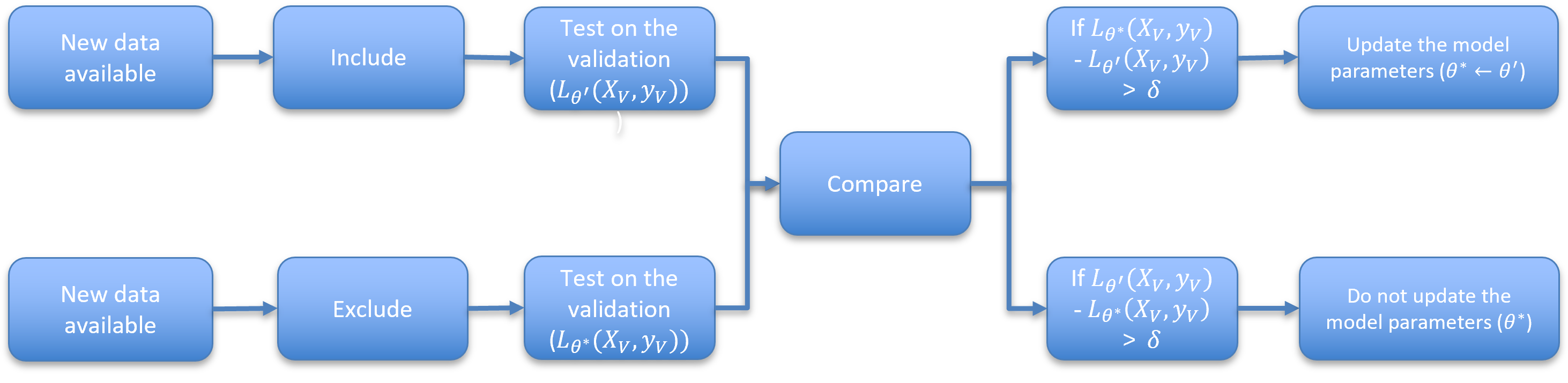}
 \caption{Adaptive online learning framework}
 \label{fig:adaptive_framework}
\end{figure*}

\noindent
\textbf{V. Selective Model Update}: The model parameters are updated only if the proposed parameters yield significantly better performance on the validation set, as determined by a predefined margin:
\begin{equation}
\text{if} \quad
L_{\bm{\theta}^*}(\bm{X}_{V}, \bm{y}_{V})
- L_{\bm{\theta}'}(\bm{X}_{V}, \bm{y}_{V})
> \delta,
\end{equation}
then the parameters are updated as
\(\bm{\theta}^* \leftarrow \bm{\theta}'\),
where \( L_{V}(\bm{\theta}^*) \) and \( L_{V}(\bm{\theta}') \) represent the validation loss for the current and proposed parameters, respectively, and \( \delta \) is a predetermined margin. This margin ensures that only meaningful improvements lead to parameter updates, preventing the model from adapting to insignificant changes that could result in overfitting or instability (by default $\delta $ can be set to zero). This approach ensures that the model is updated only when the new parameter set provides a better estimation for the validation set, maintaining robustness and improving performance incrementally. It should be pointed out that the validation set can also be dynamically updated similarly to how the model parameters are updated. This process ensures that the validation set remains relevant and reflects the most recent data distribution. As a potential extension of the model update criteria, more advanced decision-making strategies can be implemented to ensure only meaningful updates are applied. This includes using an adaptive threshold for $\delta$ that adjusts based on the training stage, updating the validation set over time, considering multiple validation metrics (e.g., loss, variance) to evaluate the model’s performance comprehensively, and incorporating confidence intervals to account for noise and variability in the validation results. By combining these techniques, the model update becomes more robust, avoiding overfitting and ensuring that updates lead to genuinely improved generalization and performance across multiple dimensions. The visualization of the framework can be seen in Figure \ref{fig:adaptive_framework}.
The online learning approach allows the model to be updated incrementally as new data becomes available. This approach allows the model to remain up-to-date with the latest trends and patterns in the data, which is particularly crucial in time series forecasting scenarios such as electricity price prediction. The process involves evaluating the model's performance incrementally and deciding whether to incorporate the new data based on its impact on the model's accuracy. The steps of the adaptive online learning model are summarized in Algorithm 1:

\begin{algorithm}[H]
\resizebox{\columnwidth}{!}{%
\begin{minipage}{\columnwidth}
\caption{Adaptive Online Learning Framework}
\begin{algorithmic}[1]
\State \textbf{Model Inputs:} Feature and output vectors for Training set $S$, Validation set $V$, New data batch set $B$, Current trained model with parameter set $\bm{\theta}^*$, and Hyperparameters (e.g., $\delta$).
\For{each new batch $b \in B$}
    \State Extract features $\bm{X}_b$ and target values $\bm{y}_b$.
    \State Estimate target values $\hat{\bm{y}}_b$ for all $b \in B$ from the prediction model with parameter $\bm{\theta}^*$.
    \State Evaluate the loss function for all points in the new dataset $B$.
    \State Update model parameters from Eq.~\eqref{eq:gra} to find $\bm{\theta}'$.
    \State Freeze certain layers (e.g., early layers) in $\bm{\theta}'$ from $\bm{\theta}^*$ to retain pre-learned features.
    \State Evaluate the loss function for all points in the validation dataset $V$ based on both $\bm{\theta}^*$ and $\bm{\theta}'$.
    \If{model performance on $V$ improves with using $B$ for training, that is,
        \[
        \text{if } L_{\bm{\theta}^*}(\bm{X}_{V}, \bm{y}_{V}) - L_{\bm{\theta}'}(\bm{X}_{V}, \bm{y}_{V}) > \delta,
        \]}
        \State Update model parameters incrementally as:
        \[
        \bm{\theta}^* \leftarrow \bm{\theta}'.
        \]
    \EndIf
    \State Discard $B$ if not relevant.
\EndFor
\State \textbf{Model Outputs:} Return the updated model with parameter set $\bm{\theta}^*$.
\end{algorithmic}
\end{minipage}%
}
\end{algorithm}

\section{Experimental Results}

\subsection{Data Description}

This article leverages a diverse array of data sources to forecast hourly electricity prices accurately. The original dataset is provided by the IISE ESD/QCRE/PG\&E Energy Analytics Challenge and includes historical data spanning from 2020 to 2023 and almost half of 2024. A preliminary version of this work by the authors received the Winner Award in this Challenge Competition. The dataset includes hourly electricity prices, load demands, and gas prices for the pricing node of NP-15 situated in Northern California, which serves as a crucial geographic location for pricing electricity within the California Independent System Operator (CAISO) market \citep{IISE2024}. Participants in the CAISO day-ahead market can submit bids for either purchasing electricity or selling generated power. Accurate forecasts of demand and prices are essential for successful market participation. The objective is to design a model that can run using the most recent data to predict hourly prices for the following day. This dataset forms the foundation of our training and testing data for developing the forecasting model. External datasets (not provided in the original study) used in this study are discussed as follows:

\noindent
- \textit{Weather Data (World Weather Online):} This dataset incorporates detailed meteorological conditions across California, such as temperature, precipitation, humidity, and wind speed. This dataset is essential for understanding the impact of weather on electricity demand and renewable energy production as shown in studies by \citet{neumann2023using}, \citet{son2017short}, and \citet{wang2019daily}. Features in this dataset are Date and Time, Maximum Temperature (°C), Minimum Temperature (°C), Total Snowfall (cm), Sunlight Hours, UV Index, Moon Illumination (\%), Moonrise Time, Moonset Time, Sunrise Time, Sunset Time, Dew Point (°C), Feels Like Temperature (°C), Heat Index (°C), Wind Chill (°C), Wind Gust Speed (km/h), Cloud Cover (\%), Humidity (\%), Precipitation (mm), Atmospheric Pressure (mb), Temperature (°C), Visibility (km), Wind Direction (°), and Wind Speed (km/h).

\noindent
- \textit{Grid Status Data (CAISO):} This dataset offers insights into the energy generation mix, including both renewable sources and conventional power generation, which has been shown to positively impact predictions, as demonstrated in the study by \citet{tschora2022electricity}. This data helps in predicting price fluctuations based on supply dynamics (e.g., Solar, Wind, Geothermal, Biomass, Biogas, Small Hydro, Coal, Nuclear, Natural Gas, Large Hydro, Batteries, and Imports). These datasets have been meticulously preprocessed and integrated into our analysis to ensure a comprehensive approach to forecasting electricity prices.

\subsection{Experimental Setup and Implementation Details}

The problem is formulated as a 24-hour-ahead vector prediction task. The proposed LSTM-based model outputs a 24-dimensional price vector using an input sequence of 72 historical time steps combined with same-day known features, resulting in a total input length of 96 time steps per sample. The network architecture consists of two stacked LSTM layers with 256 hidden units each, followed by dropout layers with a dropout rate of 0.3, and a fully connected linear output layer.

Model training is performed using the Adam optimizer with a fixed learning rate of 0.001, a batch size of 64, and a maximum of 20 epochs. Early stopping is applied based on validation MAE, with a patience of 5 epochs. A custom loss function combining MAE, JSD, and a smoothness penalty is employed, with weights of 0.001 and 0.03 assigned to the divergence and smoothness terms, respectively.

The dataset is split chronologically to prevent information leakage, with data prior to January~1,~2024 used for initial training. Model evaluation is conducted using a rolling-origin forecasting strategy, where 24-hour-ahead predictions are generated sequentially over the test period. Hyperparameters are selected based on validation performance and held fixed during testing. For the adaptive online learning framework, model updates are evaluated once per day as each new 24-hour batch becomes available. The validation set used for each update decision consists of one day, corresponding to 24 hourly observations. The update threshold is set to $\delta = 0$, meaning that a proposed update is accepted only if it reduces the validation loss relative to the current model. The validation day is updated sequentially across the test horizon so that the update decision remains temporally close to the current forecasting period while avoiding the use of future information. During each adaptive online update, the earlier LSTM layers are frozen and only the final dense output layer is updated. This strategy preserves the temporal representations learned from the historical training data while allowing the model to make limited adjustments based on newly arriving data. Restricting the update to the final layer also reduces the risk of extreme parameter fluctuations and overfitting to short-term noise, which is particularly important when the validation window consists of one day. Over the test horizon, 167 proposed updates are evaluated, of which 91 are accepted and 76 are rejected.

\noindent
\subsection{Data Preprocessing}
A critical step in our methodology involved data preprocessing and cleaning to ensure model accuracy and robustness. This process included handling missing values, correcting outliers, ensuring data consistency, and feature engineering.

\noindent
\textit{\textbf{- Handling Anomalous and Missing Data:}} Upon inspection, we identified anomalies within the 'HOUR\_ENDING' column, where some entries erroneously showed a 25th hour, alongside the absence of the 24th hour in three specific days. To address these discrepancies, entries indicating a 25th hour were considered erroneous and subsequently removed from the dataset to maintain temporal accuracy. For days missing the 24th hour data points, we employed a strategy to fill these gaps by averaging the data from the hours immediately before and after the missing entry. This approach ensured a smooth and logical transition in our time series data, preserving the integrity and continuity of our dataset.

\noindent
\textbf{\textit{- Creating Test Set \& Initial Sequence Requirement:}} A unique challenge in time series forecasting, particularly with models that rely on sequences of historical data, is making predictions at the start of the test set where preceding historical data is required but not available within the set itself. This challenge is pronounced in our LSTM model, which requires a sequence of historical data equal to the defined time steps to make accurate predictions. To overcome this challenge and ensure our model could make predictions for the entire test set, including its initial segment, we implemented a strategy that leverages historical data from the training set. Specifically, for the initial predictions in the test set where the required historical sequence falls outside the test time frame, we extended the sequence by incorporating the last rows of the training data to complete the necessary input sequence length.

\noindent \textbf{\textit{- Feature Selection:}} The robustness of our prediction model is underpinned by the careful selection and preparation of features, which include operational and market dynamics (features like actual and predicted loading MW from CAISO, PG\&E, SCE, and SDG\&E), alongside natural gas prices providing a comprehensive view of the energy market's supply-demand balance and operational costs. We also incorporated weather-related variables, such as HeatIndexC, wind direction and speed, and sun hours to capture the impact of environmental conditions on energy production and demand. Data on renewable energy production, including solar, wind, geothermal, biomass, biogas, and small hydro, are factored in to reflect the growing impact of green energy sources on electricity prices.

\noindent
\textit{\textbf{Temporal Features:}} To capture the cyclical nature of electricity demand and pricing, we engineered features based on time, such as sinusoidal transformations of hours, days of the week, and months, alongside indicators for weekends. These features serve as the foundation for our LSTM model, ensuring a comprehensive analysis that accounts for the multifaceted influences on electricity prices. We also conducted other basic feature engineering tasks, such as one-hot encoding, normalization, and handling missing values, to prepare the data for our analysis.




\subsection{Discussion on the Results of the Numerical Experiments}

Throughout this article, we discuss the results based on three separate models:

\noindent
\textbf{\textit{I. Static Model}}: In this model, the parameters are estimated using a fixed training dataset and remain unchanged for the entire duration of the test set.

\noindent
\textbf{\textit{II. Dynamic Model}}: In this model, the parameters are updated from scratch after each new batch (in this case, 1 day) of the test set becomes available. In this model, we assume that the users keep retraining the model as new data becomes available.

\noindent
\textbf{\textit{III. Adaptive Online Learning}}: This model selectively updates the parameters of the static model as new data in the test set becomes available (as discussed in Section 4). 

\begin{figure*}[!t]
 \centering
 \includegraphics[width=0.82\textwidth]{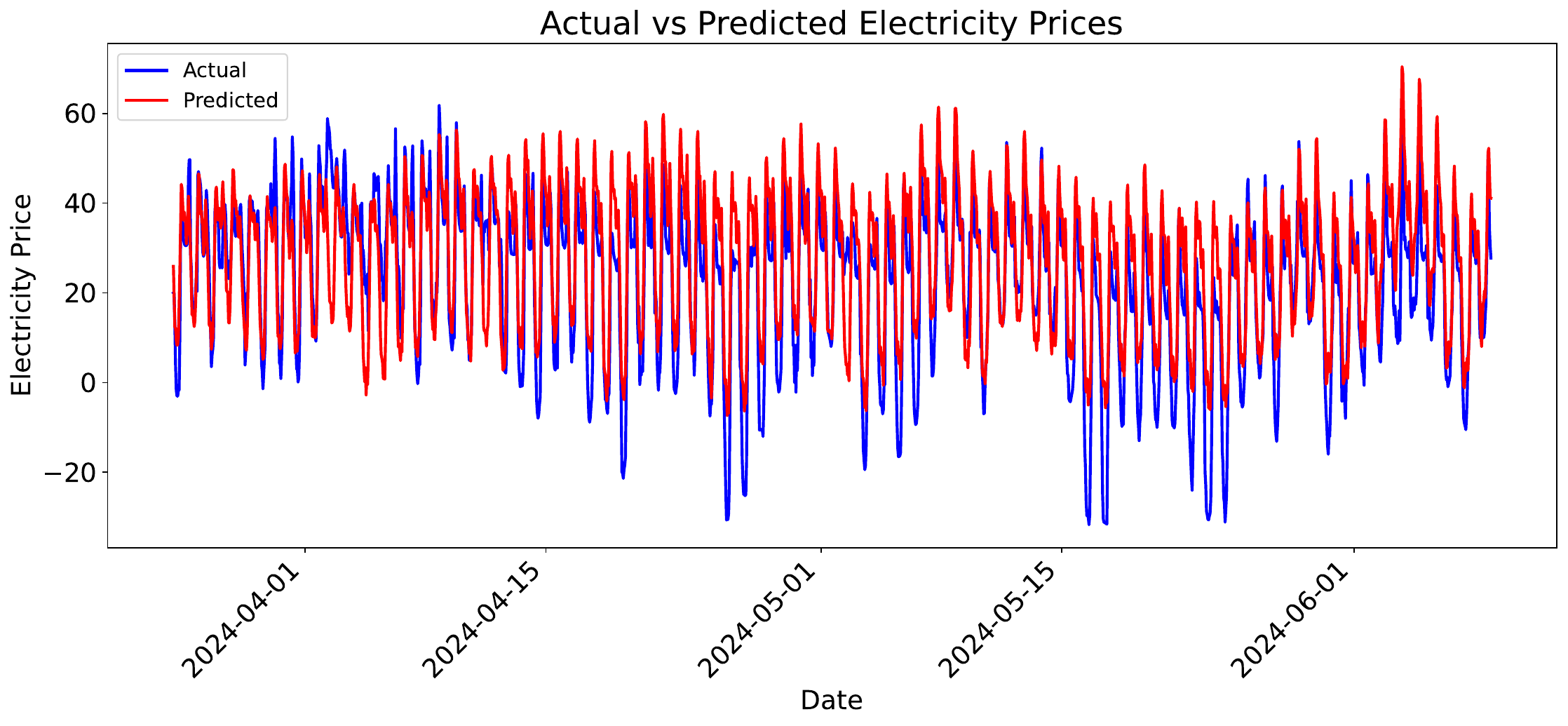}
 \caption{Prediction results for the last 3 months of the test set: the static model}
 \label{fig:resultinitial}
\end{figure*}

It should be noted that in all three models, the features used for the prediction of a specific date are the same. The inclusion of new data pertains to the labels of the data for model training. In addition to comparing the three models mentioned above, we present the impact of the custom loss function, the performance of the online learning model, the influence of energy generation mix data, and the effect of including the day-of-prediction features. For all of our numerical experiments, results are evaluated based on comparing the predicted values versus true values. A sample of predicted prices for a 3-month period based on the static model is shown in Fig. \ref{fig:resultinitial}.

\subsubsection{Overall Prediction Performance}
Table \ref{tab:comparisonloss} presents the testing performance of the static model under different loss-function configurations, including MSE, MAE, MAE + JSD, MAE + smoothness, and the full custom loss function proposed in Section 3.2. Performance is evaluated using MSE, MAE, and RMSE over the entire test set, which spans approximately six months. MSE measures the average squared prediction error, MAE measures the average absolute prediction error, and RMSE expresses the error magnitude in the same units as the predicted electricity prices. Lower values for all three metrics indicate better predictive performance.

\begin{figure*}[!t]
 \centering

 \begin{subfigure}[b]{0.48\textwidth}
 \centering
 \includegraphics[width=\textwidth]{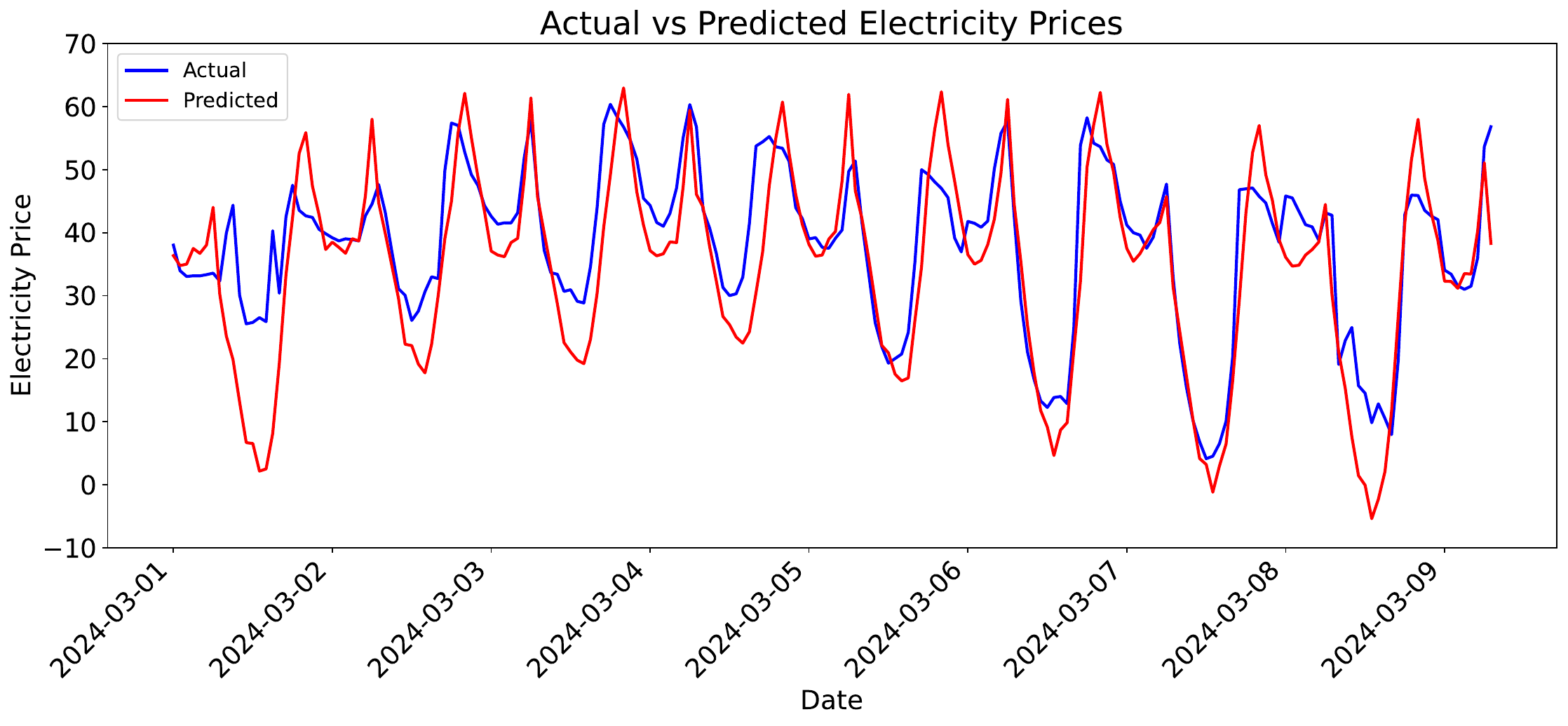}

 \vspace{0.5em}

 \includegraphics[width=\textwidth]{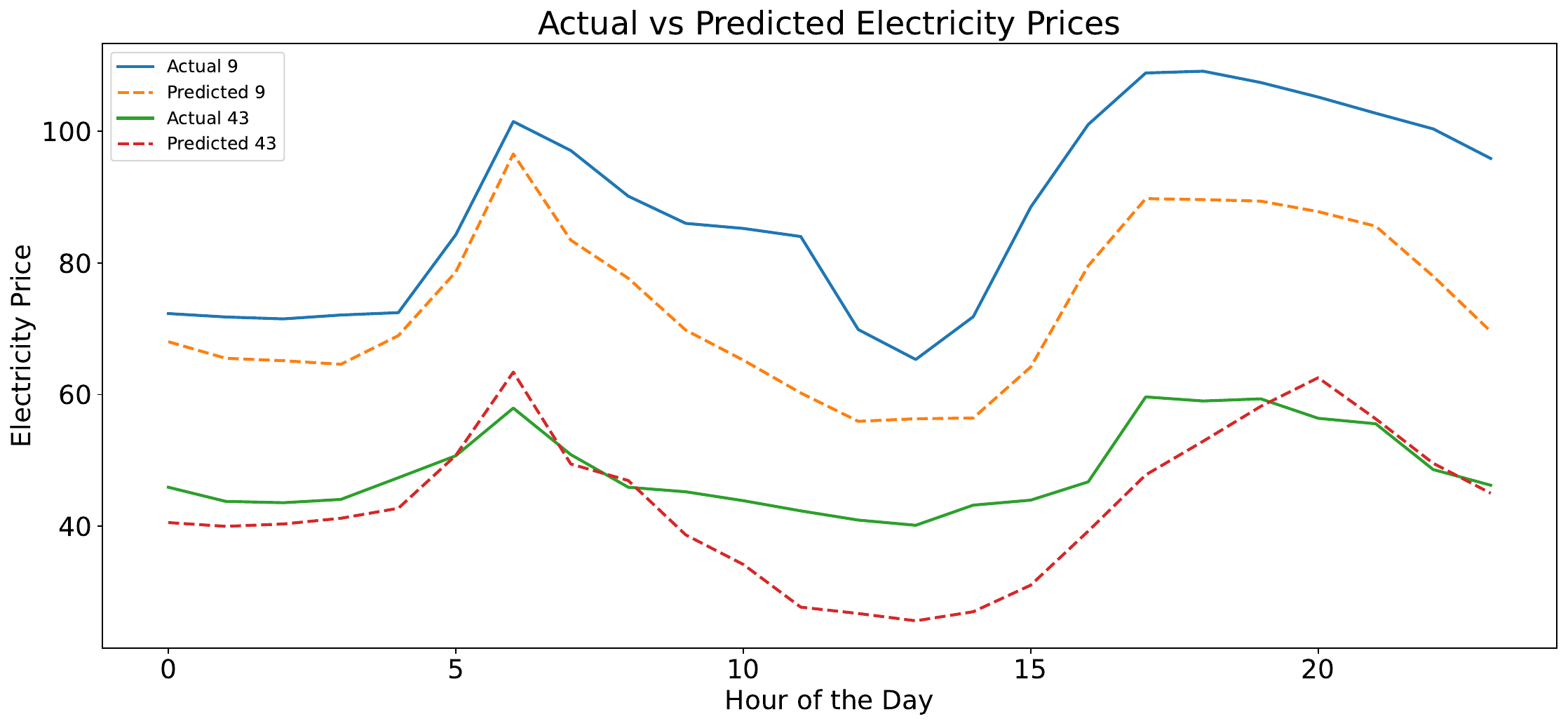}
 \caption{Before custom loss}
 \label{fig:beforereg}
 \end{subfigure}
 \hfill
 \begin{subfigure}[b]{0.48\textwidth}
 \centering
 \includegraphics[width=\textwidth]{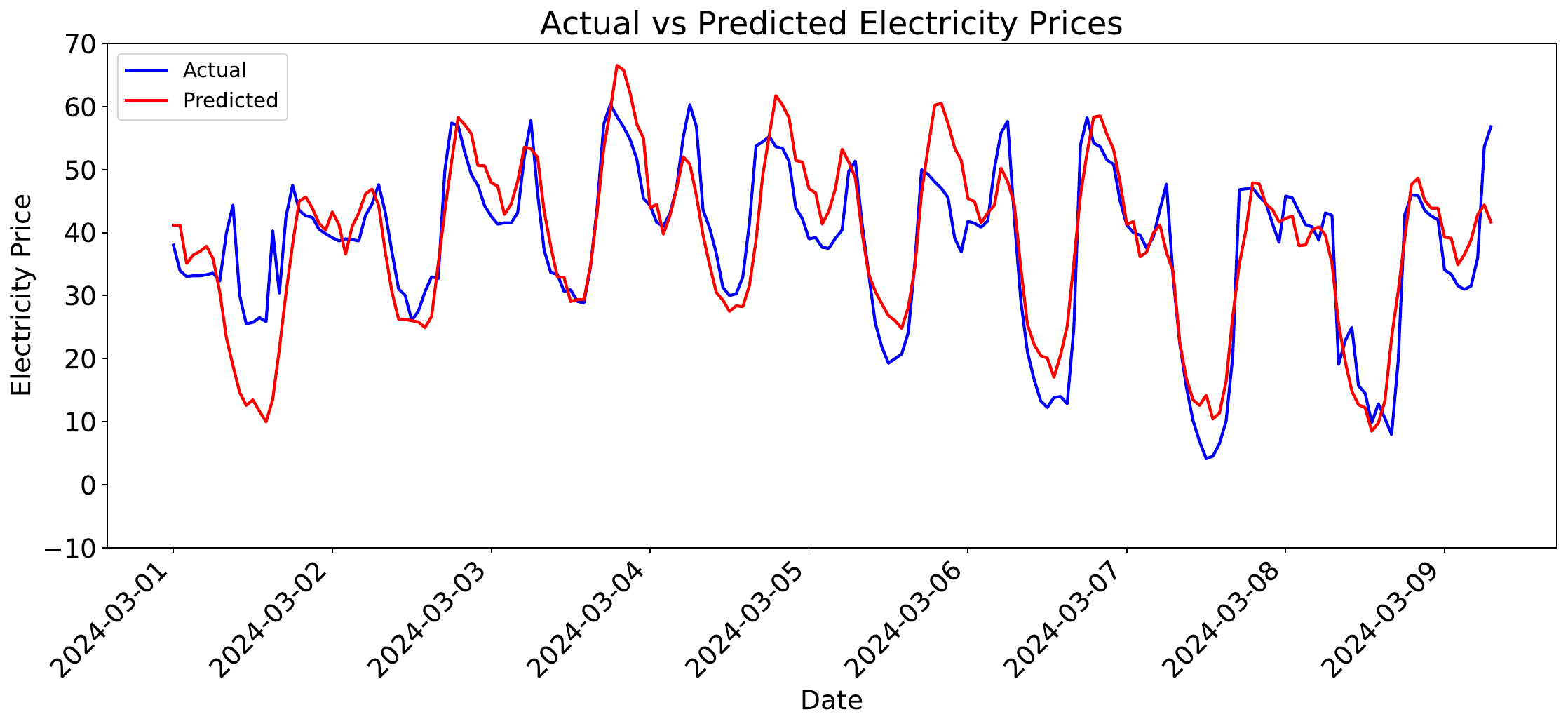}

 \vspace{0.5em}

 \includegraphics[width=\textwidth]{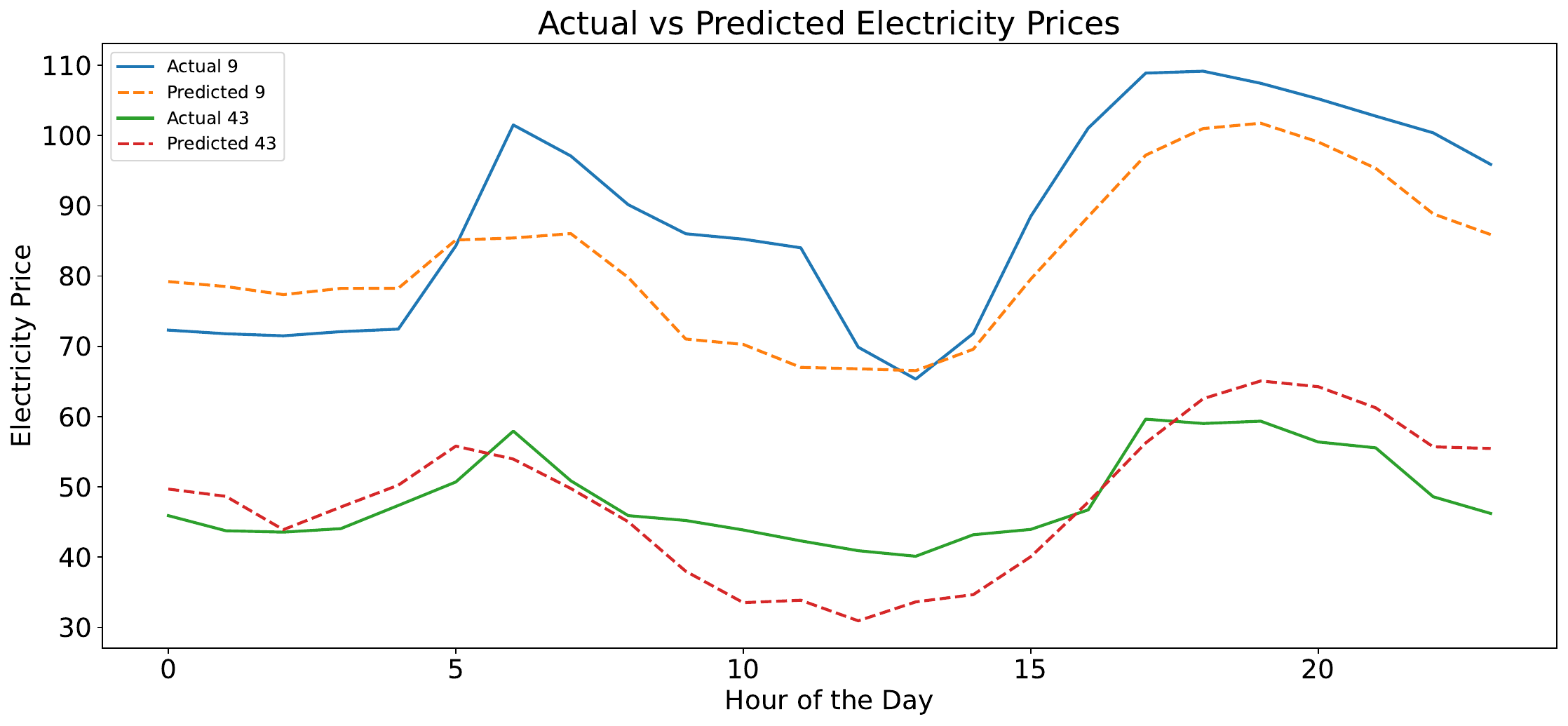}
 \caption{After custom loss}
 \label{fig:afterreg}
 \end{subfigure}

 \caption{Comparison of actual and predicted electricity prices before and after applying the custom loss function. The top row shows overall time-series predictions, while the bottom row highlights hourly predictions for two specific days.}
 \label{fig:customloss}
\end{figure*}

The results show that the full custom loss function achieves the lowest error across all metrics. The ablation study further demonstrates the individual contribution of each component. Adding JSD to MAE improves the alignment between the predicted and actual intra-day price profiles, while adding the smoothness penalty produces a larger reduction in error by discouraging abrupt hour-to-hour fluctuations. Their combination provides the best overall performance, indicating that the two regularization terms offer complementary benefits rather than redundant effects. The weights $\alpha$ and $\beta$ are tuned hyperparameters, with $\alpha = 0.001$ and $\beta = 0.03$ providing the best validation performance among the tested configurations. These relatively small values are used because the JSD and smoothness components operate on different numerical scales than the MAE term and are intended to act as regularizers rather than dominate the pointwise prediction error. Therefore, they should not be interpreted as universal constants and should be re-tuned when applying the framework to different datasets. Figure \ref{fig:customloss} compares the predicted and actual electricity prices before and after applying the custom loss function. The model using the custom loss aligns more closely with actual prices, particularly during peak intervals, and produces smoother and more reliable hourly predictions.

\begin{table}[!t]
\centering
\small
\caption{Results comparison and ablation study for different loss functions}
\label{tab:comparisonloss}
\begin{tabularx}{\columnwidth}{@{}l *{3}{>{\centering\arraybackslash}X}@{}}
\toprule
\textbf{Loss function} & \textbf{MSE} & \textbf{MAE} & \textbf{RMSE} \\
\midrule
\textbf{MSE} & 315.82 & 12.68 & 17.77 \\
\textbf{MAE} & 189.65 & 9.19 & 13.77 \\
\textbf{MAE + JSD} & 169.44 & 8.96 & 13.01 \\
\textbf{MAE + SM} & 161.23 & 7.84 & 12.70 \\
\textbf{MAE + JSD + SM} & 158.76 & 7.76 & 12.60 \\
\bottomrule
\end{tabularx}
\end{table}


In another experiment, two models—a dynamic model and a static model—were compared for their ability to predict electricity prices. The dynamic model, retrained daily to make predictions one day at a time, was contrasted with a static model trained once and used for all subsequent predictions without further updates. The performance was evaluated using MSE, MAE, and RMSE. Results in Table \ref{tab:comparisonloss2} imply that the static model outperformed the dynamic model across all metrics, with lower MSE (158.76 vs. 188.45), MAE (7.76 vs. 8.26), and RMSE (12.60 vs. 13.73). The dynamic model's higher error rates likely result from its sensitivity to daily fluctuations and noise due to constant retraining, which may lead to overfitting and making the model too sensitive to short-term fluctuations and noise rather than capturing the underlying patterns in the data. In contrast, the static model's stability and consistent understanding of the data allow it to generalize better, leading to improved prediction accuracy. The static model's lower error metrics suggest that, in this case, a well-trained model on a comprehensive dataset may outperform a model that is frequently updated but lacks stability.
\begin{table}[!t]
\centering
\small
\caption{Comparison of prediction error metrics for dynamic and static models}
\label{tab:comparisonloss2}
\begin{tabularx}{\columnwidth}{@{}Xccc@{}}
\toprule
\textbf{Model} & \textbf{MSE} & \textbf{MAE} & \textbf{RMSE} \\
\midrule
Dynamic Model & 188.45 & 8.26 & 13.73 \\
Static Model  & 158.76 & 7.76 & 12.60 \\
\bottomrule
\end{tabularx}
\end{table}
\subsubsection{Results of the Adaptive Online Learning Model}
\begin{figure*}[!t]
 \centering

 \begin{subfigure}[b]{0.48\textwidth}
 \centering
 \includegraphics[width=\textwidth]{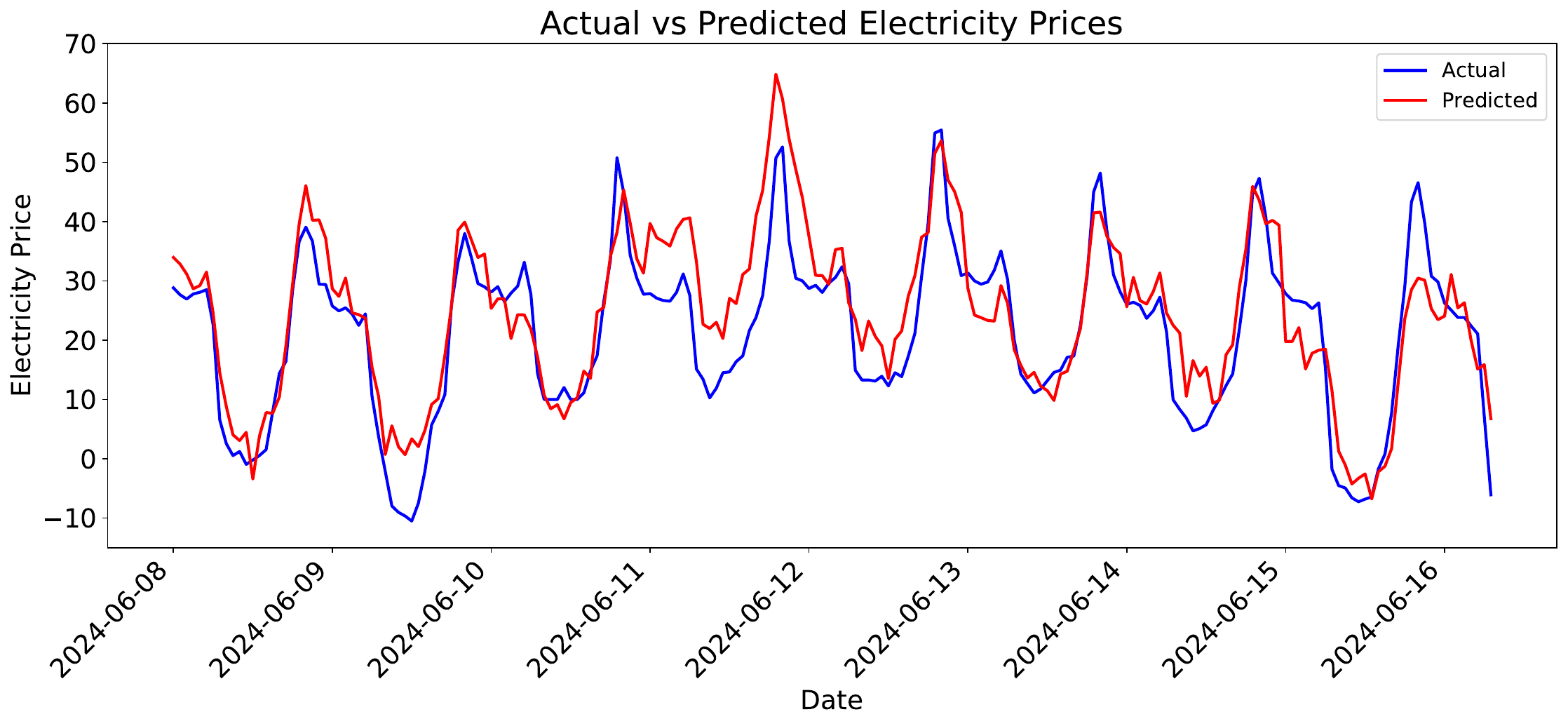}

 \vspace{0.5em}

 \includegraphics[width=\textwidth]{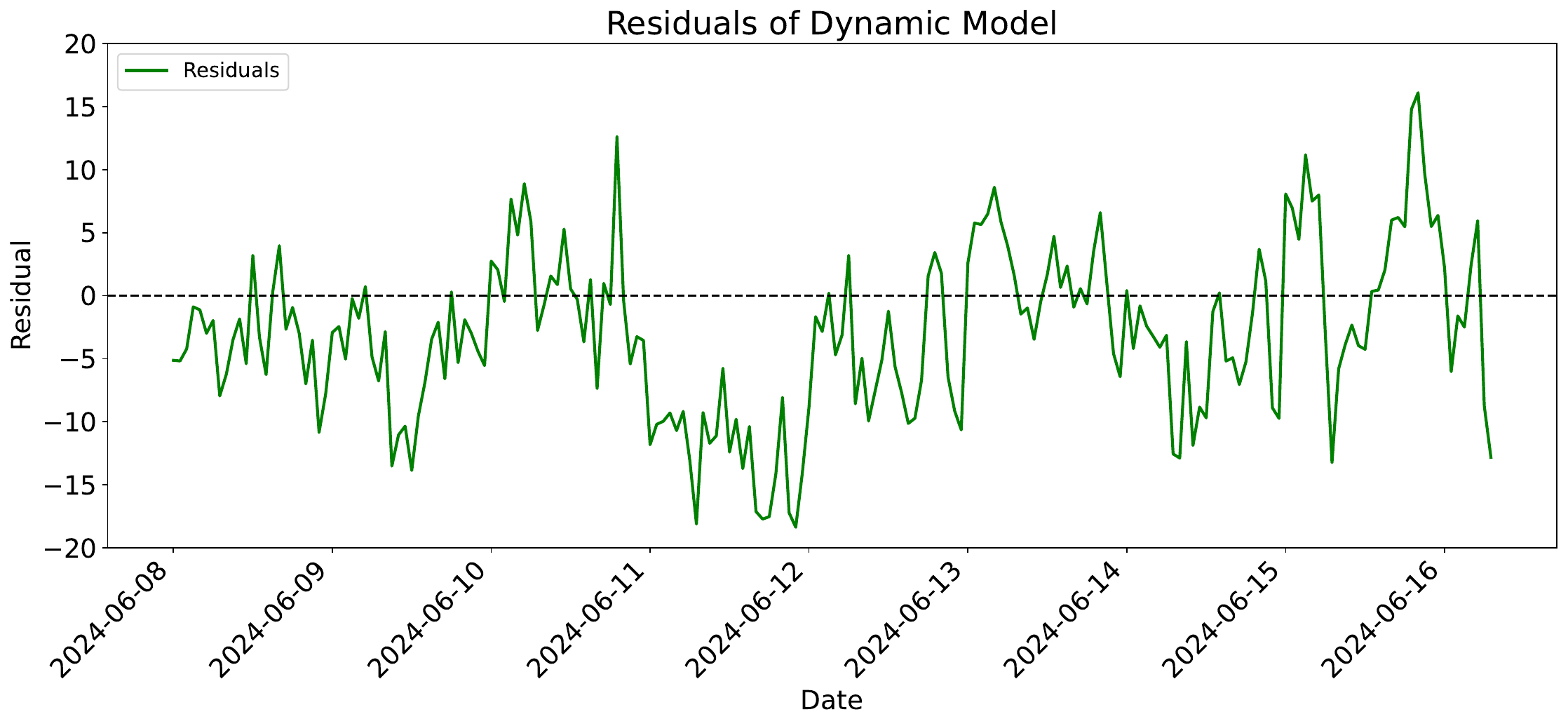}
 \caption{Dynamic model}
 \label{fig:dynamic_model}
 \end{subfigure}
 \hfill
 \begin{subfigure}[b]{0.48\textwidth}
 \centering
 \includegraphics[width=\textwidth]{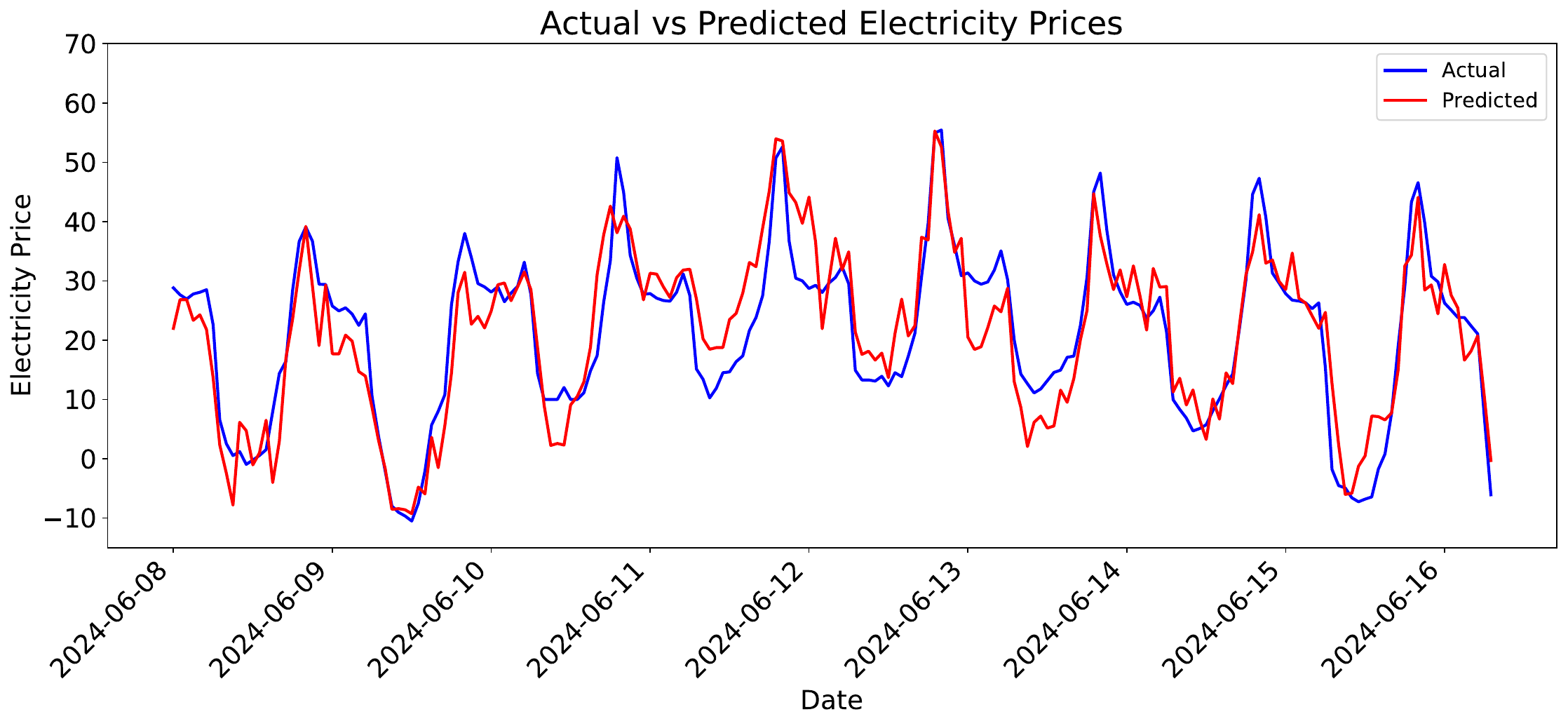}

 \vspace{0.5em}

 \includegraphics[width=\textwidth]{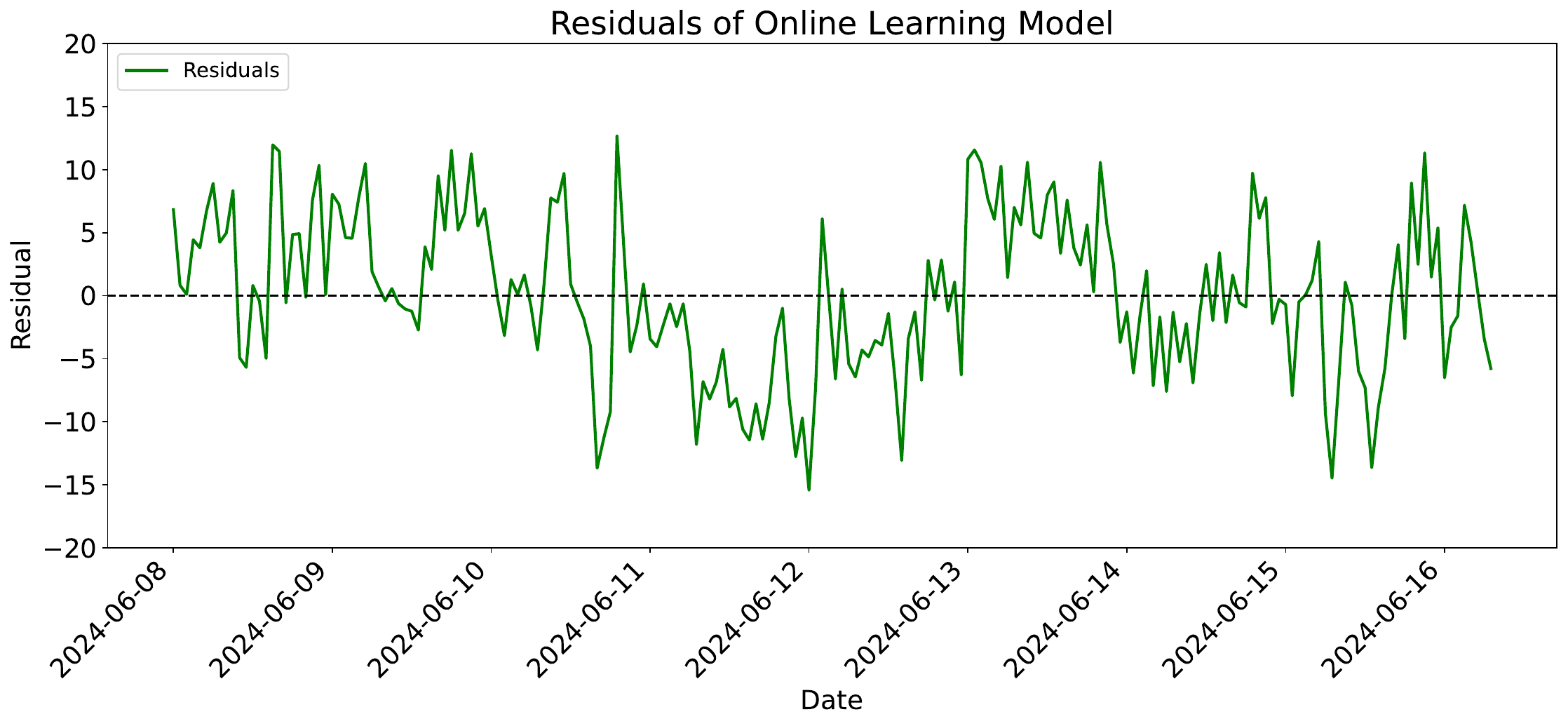}
 \caption{Online learning model}
 \label{fig:online_learning_model}
 \end{subfigure}

 \caption{Results and residuals comparison between dynamic and online learning models}
 \label{fig:results_and_residuals_comparison}
\end{figure*}

In this experiment, the performance of three models—Online Learning, Dynamic, and Static—was compared in the context of electricity price forecasting. The Online Learning Model continuously updates its parameters daily using the most recent data, adapting to new trends and maintaining relevance in its predictions. For each new batch of data, the model's predictions are evaluated using a designated test period immediately following the data batch, ensuring a forward-looking assessment. The Dynamic Model, on the other hand, is retrained daily but without selective updates, relying solely on the most recent data. The Static Model is trained once and does not update with new data, making predictions based on the initial training data throughout the experiment. Results shown in Table \ref{tab:performance_comparison_online} indicate that the Online Learning Model outperforms both the Dynamic and Static models. Specifically, the Online Learning Model achieves the lowest error metrics, indicating its superior predictive accuracy. The Dynamic Model, despite being retrained daily, shows higher errors, likely due to its sensitivity to daily fluctuations without leveraging previous trends effectively. The Static Model performs better than the Dynamic Model but still lags behind the Online Learning Model, with moderate error metrics. The superior performance of the Online Learning Model can be attributed to its ability to incrementally incorporate new data batches while maintaining the context of previous information. This iterative updating mechanism allows the model to adapt effectively to evolving trends and new information, resulting in better predictive accuracy and generalization capabilities. Overall, these results underscore the efficacy of online learning in dynamic environments, offering a practical advantage for real-time predictive tasks like electricity price forecasting.
\begin{table}[!t]
\centering
\small
\caption{Performance comparison between online, dynamic, and static models}
\label{tab:performance_comparison_online}
\begin{tabularx}{\columnwidth}{@{}Xccc@{}}
\toprule
\textbf{Model} & \textbf{MSE} & \textbf{MAE} & \textbf{RMSE} \\
\midrule
Online Learning Model & 122.25 & 7.50 & 11.06 \\
Dynamic Model & 188.45 & 8.26 & 13.73 \\
Static Model & 158.76 & 7.76 & 12.60 \\
\bottomrule
\end{tabularx}
\end{table}
Figure \ref{fig:results_and_residuals_comparison} provides a visual comparison between the Dynamic Model and the Online Learning Model, focusing on their predicted electricity prices and the associated residuals. The top row of plots shows the actual versus predicted electricity prices for both models. While both models capture the general trend, the Online Learning Model (right) shows predictions that more closely align with the actual values compared to the Dynamic Model (left). This suggests that the Online Learning Model is better at adapting to the evolving patterns in the data. The bottom row provides a residual analysis, where the residuals represent the difference between the actual and predicted values. The residuals for the Online Learning Model (bottom right) exhibit less variability and are more centered around zero compared to the Dynamic Model (bottom left). This reduced variability indicates that the Online Learning Model is not only more accurate but also more consistent in its predictions. 

\subsubsection{Impact of Energy Generation Mix Dataset}
Incorporating the energy generation mix dataset into our models has a profound effect on improving predictive accuracy, as demonstrated by the significant reduction in key error metrics. The detailed data regarding the types and proportions of energy sources used in electricity generation provides crucial insights into the supply-side conditions, enabling the model to more effectively capture the underlying trends and fluctuations in electricity pricing. This enriched dataset allows the models to produce more precise and reliable forecasts by better understanding the dynamics of energy supply. As shown in Table \ref{tab:performance_comparison_fuel}, models that include fuel mix data consistently exhibit lower MSE, MAE, and RMSE compared to those that do not incorporate this information. For instance, the Static Model with fuel mix data shows a marked improvement over the Static Model without it, reducing MSE from 252.27 to 158.76. Similarly, the Online Learning Model with fuel mix data also outperforms its counterpart without the fuel mix, with MSE dropping from 230.38 to 122.25. These results highlight the importance of incorporating detailed energy generation data in predictive modeling, as it leads to significantly enhanced performance across different modeling approaches.
\begin{table}[!t]
\centering
\small
\caption{Impact of fuel mix data on predictive performance across different models}
\label{tab:performance_comparison_fuel}
\begin{tabularx}{\columnwidth}{@{}Xccc@{}}
\toprule
\textbf{Model} & \textbf{MSE} & \textbf{MAE} & \textbf{RMSE} \\
\midrule
Static Model-FM     & 158.76 & 7.76  & 12.60 \\
Static Model-NFM    & 252.27 & 10.15 & 15.88 \\
Online Learning-FM  & 122.25 & 7.50  & 11.06 \\
Online Learning-NFM & 230.38 & 9.28  & 15.18 \\
\midrule
\multicolumn{4}{@{}p{\columnwidth}@{}}{\footnotesize
FM denotes the model with fuel mix data; NFM denotes the model without fuel mix data.} \\
\bottomrule
\end{tabularx}
\end{table}

\subsubsection{Impact of Including Same-Day Features in Predictions}
In the original IISE ESD/QCRE/PG\&E Energy Analytics Challenge setting, same-day features were permitted as inputs for generating day-ahead price predictions. This reflects the real-world operational setting considered by the company, where forecasts or operationally available same-day information can be used to support day-ahead forecasting decisions.  Accordingly, in this study, same-day inputs are treated as forecast-available or operationally available variables at the prediction time, rather than ex-post realized values observed after the target day. This clarification is important to distinguish the use of prediction-time-available information from potential data leakage.

Incorporating same-day features into the prediction model significantly enhances short-term accuracy by leveraging the most up-to-date information available. This enables the model to more effectively capture immediate trends, fluctuations, and events that impact electricity prices, such as weather changes, demand spikes, or supply disruptions. The ability to include same-day data allows the model to adapt in real-time, making it highly responsive to dynamic conditions. This real-time adaptability is crucial for applications where immediate responsiveness is required, ensuring the model can promptly adjust to sudden market changes, thereby enhancing overall performance. Moreover, utilizing same-day features enables the model to maintain a comprehensive temporal context, balancing both historical trends and current market conditions. This enriched temporal context results in more informed and accurate predictions, as evidenced by the significant improvements in key error metrics. 
\begin{table}[!t]
\centering
\small
\caption{Impact of day-of-prediction features on different models}
\label{tab:performance_comparison_current_day}
\begin{tabularx}{\columnwidth}{@{}Xccc@{}}
\toprule
\textbf{Model} & \textbf{MSE} & \textbf{MAE} & \textbf{RMSE} \\
\midrule
Dynamic Model-CD     & 188.45 & 8.26  & 13.73 \\
Dynamic Model-NCD    & 328.51 & 10.54 & 18.12 \\
Online Learning-CD      & 122.25 & 7.50  & 11.06 \\
Online Learning-NCD     & 286.42 & 9.64  & 16.92 \\
\midrule
\multicolumn{4}{@{}p{\columnwidth}@{}}{\footnotesize
CD denotes the model with current-day data; NCD denotes the model without current-day data.} \\
\bottomrule
\end{tabularx}
\end{table}
For instance, as shown in Table \ref{tab:performance_comparison_current_day}, the Dynamic Model's MSE decreases from 328.51 to 188.45 when same-day data is included. Similarly, the Online Learning Model shows a marked reduction in MSE from 286.42 to 122.25 when incorporating same-day features. These results highlight the importance of including current-day feature data in the predictive modeling process, as it leads to substantially improved accuracy and reliability in forecasting electricity prices.

\subsubsection{Comparison with Additional Benchmark Models}
While all our numerical experiments discussed earlier compare our model with baseline and similar approaches, this section focuses on comparing various machine learning models applied to electricity price prediction. For this experiments, we use the same training and testing sets for each model. These benchmark models are the Support Vector Regressor (SVR), Decision Tree Regressor (DTR), Gradient Boosting Regressor (GBR), Autoregressive Integrated Moving Average (ARIMA), and Seasonal Autoregressive Integrated Moving Average (SARIMA). SVR is a robust model for capturing linear dependencies but may underperform when faced with the inherent non-linearity of electricity price data. DTR, which splits data into subsets using decision rules, can struggle with complex temporal patterns. GBR improves performance through boosting weak models, offering better results than simpler models \citep{zhou2019optimized}. Finally, the ARIMA and SARIMA models, commonly used for time series forecasting, focuses on autoregressive and moving average components but has limitations when dealing with rapidly changing variables in energy markets [\citealp{jiang2018day, wang2023short}]. Multivariate Autoregressive Integrated Moving Average (MARIMA) extends SARIMA by incorporating exogenous variables, enabling it to account for external drivers such as weather and generation mix. As shown in Table \ref{tab:performance_comparison_other}, the online learning approach outperforms these traditional models. All machine learning benchmarks are trained using the same input features and training-testing splits as the proposed model to ensure a fair comparison. For ARIMA-family models, standard autoregressive formulations are used, with ARIMA and SARIMA implemented as univariate models and MARIMA incorporating the same exogenous variables as the proposed framework. Hyperparameters for benchmark models are selected based on validation performance or standard library defaults when appropriate, and all models are evaluated using the same rolling-origin forecasting strategy and error metrics. 

Compared to classical time-series benchmarks such as ARIMA, SARIMA, and MARIMA, the proposed framework benefits from its ability to model nonlinear temporal dependencies and avoid excessive mean reversion, which is particularly important under volatile electricity price dynamics. It is also noted that multi-step forecasts produced by ARIMA-family models tend to converge toward the mean of the stationary process as the forecast horizon increases, which may limit their ability to capture intraday price variability over longer horizons. While MARIMA incorporates exogenous variables, its linear structure limits its capacity to capture complex interactions between historical prices and external drivers. Relative to conventional machine learning benchmarks, including SVR, DTR, and GBR, the proposed approach differs in both temporal modeling and learning strategy. These benchmark models are trained in a static manner and primarily optimize pointwise prediction accuracy, which restricts their ability to adapt to evolving market conditions and to enforce coherence across the forecast horizon. In contrast, the proposed adaptive online learning framework integrates an LSTM-based temporal model with a custom loss function and a selective update mechanism. This combination enables the model to continuously adapt to nonstationary market behavior while maintaining stable intraday price trajectories, which collectively explains the superior performance observed across comparative experiments.
\begin{table}[!t]
\centering
\small
\caption{Performance comparison with additional benchmark models}
\label{tab:performance_comparison_other}
\begin{tabularx}{\columnwidth}{@{}Xccc@{}}
\toprule
\textbf{Model} & \textbf{MSE} & \textbf{MAE} & \textbf{RMSE} \\
\midrule
\textbf{Online Learning} & \textbf{122.25} & \textbf{7.50} & \textbf{11.06} \\
Dynamic Model & 188.45 & 8.26 & 13.73 \\
Static Model & 158.76 & 7.76 & 12.60 \\
SVR & 247.73 & 7.87 & 15.74 \\
DTR & 260.24 & 8.46 & 16.13 \\
GBR & 269.79 & 7.69 & 16.43 \\
ARIMA & 396.27 & 12.83 & 19.90 \\
SARIMA & 288.48 & 8.41 & 16.98 \\
MARIMA & 179.47 & 8.27 & 13.40 \\
\bottomrule
\end{tabularx}
\end{table}
\subsubsection{Statistical Significance and Robustness Analysis}
To rigorously assess the comparative performance of the proposed framework, paired statistical significance and robustness analyses are conducted. For each forecast day, hourly prediction errors are aggregated into daily MAE and RMSE, yielding one paired observation per model per day. This aggregation avoids artificial inflation of sample size due to intra-day correlation and enables fair day-level comparisons. Rather than performing exhaustive pairwise testing across all benchmarks, targeted comparisons are selected to directly evaluate the main methodological contributions and the strongest competing models. Specifically, comparisons are conducted between the online learning model and (i) the static baseline, (ii) GBR, which achieves the lowest MAE among the benchmark models, and (iii) MARIMA, which attains the lowest RMSE among the benchmark models. Additional ablation studies evaluate the impact of the proposed custom loss function and the inclusion of generation mix features.

For each comparison, paired daily error differences are computed. In Table \ref{tab:statistical_comparisons}, positive values of $\Delta$MAE and $\Delta$RMSE indicate that the proposed model or configuration achieves lower error than the corresponding benchmark or ablation setting. For example, in the Online vs Static comparison, $\Delta$MAE is computed as $\text{MAE}_{\text{Static}}-\text{MAE}_{\text{Online}}$. Statistical significance is assessed using the paired Wilcoxon signed-rank test, which evaluates whether the median of the paired differences deviates from zero without assuming normality. This test is well suited to electricity price forecasting errors, which are often heavy-tailed and skewed. To complement hypothesis testing and quantify robustness, a block bootstrap procedure with contiguous day blocks is employed to estimate 95\% confidence intervals for the mean performance improvement. By resampling consecutive days rather than individual observations, this approach preserves temporal dependence in forecasting errors and provides a conservative assessment of improvement stability over time. Since multiple targeted comparisons are conducted, the results are interpreted cautiously, with particular emphasis on comparisons whose confidence intervals do not include zero.

The paired Wilcoxon signed-rank tests indicate reductions in daily MAE for the online learning model relative to the static baseline, GBR, and MARIMA. However, the bootstrap confidence intervals show that the robustness of these improvements varies across comparisons. For Online vs Static, the MAE confidence interval is strictly positive, while the RMSE interval includes zero, indicating that the MAE improvement is more stable than the RMSE improvement. For Online vs GBR, both MAE and RMSE confidence intervals include zero; therefore, although the Wilcoxon tests suggest directional daily improvements, the mean improvements are not statistically conclusive under the bootstrap confidence intervals. In contrast, the Online vs MARIMA comparison shows strictly positive confidence intervals for both MAE and RMSE, indicating more robust improvements. The ablation results also show that the inclusion of fuel mix features provides robust improvements in both MAE and RMSE. For the custom loss comparison, the RMSE improvement is robust based on the confidence interval, while the MAE confidence interval slightly includes zero and should therefore be interpreted more cautiously. All statistical comparison results are summarized in Table \ref{tab:statistical_comparisons}.

\begin{table*}[!t]
\centering
\small
\setlength{\tabcolsep}{3pt}
\caption{Comparison of online learning model and key ablations/benchmarks}
\label{tab:statistical_comparisons}
\begin{tabularx}{\textwidth}{@{}Xcccccc@{}}
\toprule
\textbf{Comparison} &
$\Delta$MAE &
$\Delta$RMSE &
$p_{\text{MAE}}$ &
$p_{\text{RMSE}}$ &
\textbf{95\% CI (MAE)} &
\textbf{95\% CI (RMSE)} \\
\midrule

Online vs Static
& 0.785
& 0.606
& 0.0316
& 0.2566
& (0.046, 1.582)
& (-0.093, 1.341) \\

Static Custom vs MAE
& 0.645
& 0.948
& $7.51{\times}10^{-4}$
& $1.41{\times}10^{-5}$
& (-0.048, 1.401)
& (0.238, 1.790) \\

Online w/o Fuel Mix
& 1.854
& 1.848
& $7.64{\times}10^{-4}$
& $3.70{\times}10^{-4}$
& (0.377, 3.664)
& (0.421, 3.618) \\

Online vs GBR
& 0.247
& 0.088
& $5.34{\times}10^{-6}$
& $1.47{\times}10^{-5}$
& (-1.404, 3.040)
& (-1.578, 2.801) \\

Online vs MARIMA
& 1.767
& 2.386
& $7.43{\times}10^{-8}$
& $1.91{\times}10^{-9}$
& (0.936, 2.653)
& (1.406, 3.417) \\

\bottomrule
\end{tabularx}
\end{table*}

\subsubsection{Computational Requirements}

In addition to predictive accuracy, computational efficiency is an important consideration for practical deployment of day-ahead electricity price forecasting models. Table \ref{tab:runtime_summary} reports the average training or update runtime per forecast day for the main models considered in this study. For models that are retrained for each forecast day, the reported value corresponds to the average time required to retrain the model and generate the next 24-hour forecast. For the adaptive online learning model, the reported value corresponds to the average time required to evaluate the new daily batch, perform the selective update when accepted, and generate the next-day forecast. The static model is trained once before the test horizon and is not retrained for each forecast day; therefore, no repeated daily runtime is reported for this model.

\begin{table}[!t]
\centering
\small
\caption{Average training/update runtime per forecast day}
\label{tab:runtime_summary}
\begin{tabularx}{\columnwidth}{@{}Xcc@{}}
\toprule
\textbf{Model} & \textbf{Strategy} & \textbf{Runtime (s)} \\
\midrule
Online Learning & Selective update & 229.17 \\
Dynamic Model & Daily retraining & 654.32 \\
Static Model & No retraining & -- \\
SVR & Daily retraining & 0.19 \\
DTR & Daily retraining & 0.27 \\
GBR & Daily retraining & 0.52 \\
ARIMA & Daily retraining & 45.69 \\
SARIMA & Daily retraining & 274.97 \\
MARIMA & Daily retraining & 1018.47 \\
\bottomrule
\end{tabularx}
\end{table}

The main drivers of computational runtime are the model structure, training strategy, update frequency, and input size. For the LSTM-based models, runtime is affected by the number of LSTM layers and hidden units, the input sequence length, the number of epochs, and whether the full network is retrained or only part of the model is updated. The dynamic model is computationally more demanding because it retrains the full network for each forecast day. In contrast, the online learning model reduces the update cost by applying selective updates and freezing the earlier LSTM layers, so that only the final dense layer is updated during online adaptation. For the benchmark models, runtime depends on the model-specific fitting procedure and the amount of historical data used at each daily retraining step. The SVR, DTR, and GBR benchmarks are substantially faster because they are implemented as tabular machine learning models trained on compact rolling feature sets, avoiding both recurrent neural network training and repeated likelihood-based time-series estimation. In contrast, ARIMA-type models require more computation due to repeated time-series refitting. In particular, MARIMA has the highest average runtime because of its multivariate structure and repeated daily refitting.

\section{Conclusion}
This research presents an advanced LSTM-based predictive model for forecasting day-ahead electricity prices, integrating key features such as historical price data, weather conditions, and the energy generation mix. The model's accuracy is enhanced by the introduction of a custom loss function combining MAE, Jensen-Shannon Divergence (JSD), and a smoothness penalty, which improves pointwise accuracy, intra-day profile alignment, and prediction stability. The adoption of an adaptive online learning approach further strengthens the model by allowing it to incorporate new data selectively and remain responsive to changing market conditions. This incremental updating process enables the model to adapt effectively to the dynamic energy market and outperform static and traditional dynamic models.

Incorporating the energy generation mix as a feature also proves important, as it provides additional information about supply-side conditions and improves the model's ability to capture price fluctuations. In addition to the LSTM-based models, several benchmark models are evaluated for electricity price forecasting. Overall, the results demonstrate the value of comprehensive feature integration, custom loss design, and adaptive learning for improving day-ahead electricity price forecasting.

The proposed forecasting framework is designed to be broadly applicable beyond the specific market and dataset considered in this study. The vector-based prediction formulation, custom loss function, and adaptive online learning strategy are not market-specific and can be extended to other electricity markets, provided that sufficient historical price data and relevant exogenous variables are available. The framework can also be adapted to alternative forecasting horizons by adjusting the output dimensionality and input window length. However, practical deployment in other markets or organizations depends on the availability, consistency, and timeliness of required operational data streams. In particular, implementation may require access to operator-level load and price data, reliable weather forecast feeds, and timely generation-mix information. These data sources may not be shared uniformly across organizations due to privacy, regulatory, commercial, or infrastructure constraints. Differences in reporting frequency, feature definitions, data latency, and market rules may also affect model transferability.

While the overall methodology is general, scenario-specific tuning of hyperparameters and model configurations may be required to accommodate differences in market structure, data quality, volatility patterns, and forecasting horizons. Model performance may degrade in the presence of abrupt structural changes, limited availability of exogenous inputs, delayed or inconsistent data streams, or shifts in market dynamics that are not adequately reflected in the training data. These limitations highlight important directions for future research. Building on this foundation, subsequent studies may incorporate additional features, further refine the custom loss function, employ dynamic hyperparameter optimization, and extend the proposed framework to different regions, markets, and forecasting horizons. These efforts will further enhance the model's robustness and generalizability, contributing valuable tools for stakeholders in the energy sector to navigate the challenges of volatility and non-linearity in electricity prices.

\section*{Acknowledgement}
The authors would like to express their sincere gratitude to Pacific Gas \& Electric Company (PG\&E) for providing the valuable data that made this study possible. We would also like to acknowledge the Energy Systems (ES) and Quality Control \& Reliability Engineering (QCRE) divisions of the Institute of Industrial and Systems Engineers (IISE) for continuous support throughout this project and for organizing the Energy Analytics Challenge. Furthermore, the authors extend their deepest appreciation to Dr. Ramin Moghaddass for his exceptional guidance, mentorship, and insightful contributions, which greatly enhanced the quality and rigor of this study. The preliminary version of this work, completed as part of the 2024 IISE PG\&E Energy Analytics Challenge, received first place in the competition. A summary of the competition and its outcomes was published in \citep{aziz2025iise}.

\section*{Declarations}

\textbf{Author contributions:} Conceptualization, S.S., I.A. and A.A.; methodology, S.S., I.A. and A.A.; software, S.S.; validation, S.S., I.A. and A.A.; formal analysis, S.S.; investigation, S.S., I.A. and A.A.; data curation, S.S. and I.A.; writing---original draft preparation, S.S.; writing---review and editing, S.S., I.A. and A.A.; visualization, S.S. All authors have read and agreed to the published version of the manuscript. \\

\noindent
\textbf{Funding:} This research received no external funding.  \\

\noindent
\textbf{Data availability:} The dataset used in this work is publicly available. \\

\noindent
\textbf{Conflict of interest:} The authors declare no conflict of interest or competing interest. \\

\noindent
\textbf{Ethics approval and consent to participate:} Not applicable.

\bibliography{sn-bibliography}

\end{document}